\documentclass{article}

\usepackage{amsmath}
\usepackage{amssymb}
\usepackage{mathtools}
\usepackage{amsthm}
\usepackage{microtype}
\usepackage{graphicx}
\usepackage{subfigure}
\usepackage{booktabs} 
\usepackage{amsfonts,amssymb}
\usepackage[inkscapelatex=false]{svg}
\usepackage{multicol}
\usepackage{multirow}
\usepackage{color}
\usepackage{threeparttable}
\usepackage{booktabs}
\usepackage{algorithmic}
\usepackage{url}

\newtheorem{myTheo}{Theorem}

\newcommand{\model}{ATFNet}

\newcommand{\boldres}[1]{{\textbf{\textcolor{red}{#1}}}}
\newcommand{\secondres}[1]{{\underline{\textcolor{blue}{#1}}}}

\newcommand{\bz}{\boldsymbol{z}}
\newcommand{\bx}{\boldsymbol{x}}
\newcommand{\by}{\boldsymbol{y}}
\newcommand{\bW}{\boldsymbol{W}}
\newcommand{\bU}{\boldsymbol{U}}
\newcommand{\bV}{\boldsymbol{V}}
\newcommand{\bE}{\boldsymbol{\epsilon}}

\usepackage{hyperref}

\renewcommand{\algorithmicrequire}{\textbf{Input:}}
\renewcommand{\algorithmicensure}{\textbf{Output:}}

\usepackage[accepted]{icml2024}

\icmltitlerunning{}

\begin{document}

\twocolumn[
\icmltitle{ATFNet: Adaptive Time-Frequency Ensembled Network for Long-term Time Series Forecasting}


\icmlsetsymbol{equal}{*}

\begin{icmlauthorlist}
\icmlauthor{Hengyu Ye}{sjtu}
\icmlauthor{Jiadong Chen}{sjtu}
\icmlauthor{Shijin Gong}{ustc}
\icmlauthor{Fuxin Jiang}{byte}
\icmlauthor{Tieying Zhang}{byte}
\icmlauthor{Jianjun Chen}{byte}
\icmlauthor{Xiaofeng Gao}{sjtu}
\end{icmlauthorlist}

\icmlaffiliation{sjtu}{Department of Computer Science and Engineering, Shanghai Jiao Tong University, Shanghai, China}
\icmlaffiliation{ustc}{School of Management, University of Science and Technology of China, Hefei, Anhui, China}
\icmlaffiliation{byte}{ByteDance Inc.}
\icmlcorrespondingauthor{Xiaofeng Gao}{gao-xf@cs.sjtu.edu.cn}

\icmlkeywords{Machine Learning, ICML}

\vskip 0.3in
]

\printAffiliationsAndNotice{}

\begin{abstract}
The intricate nature of time series data analysis benefits greatly from the distinct advantages offered by time and frequency domain representations. While the time domain is superior in representing local dependencies, particularly in non-periodic series, the frequency domain excels in capturing global dependencies,  making it ideal for series with evident periodic patterns. To capitalize on both of these strengths, we propose \model, an innovative framework that combines a time domain module and a frequency domain module to concurrently capture local and global dependencies in time series data. Specifically, we introduce Dominant Harmonic Series Energy Weighting, a novel mechanism for dynamically adjusting the weights between the two modules based on the periodicity of the input time series. In the frequency domain module, we enhance the traditional Discrete Fourier Transform (DFT) with our Extended DFT, designed to address the challenge of discrete frequency misalignment. Additionally, our Complex-valued Spectrum Attention mechanism offers a novel approach to discern the intricate relationships between different frequency combinations. Extensive experiments across multiple real-world datasets demonstrate that our \model\ framework outperforms current state-of-the-art methods in long-term time series forecasting. Codes are publicly available at  \href{https://github.com/YHYHYHYHYHY/ATFNet}{https://github.com/YHYHYHYHYHY/ATFNet}. 
\end{abstract}

\section{Introduction}
\label{introduction}

\begin{figure}[h]
\centering
\subfigure[ETTh1  Variable: ‘OT’]{
\includegraphics[width=3in]{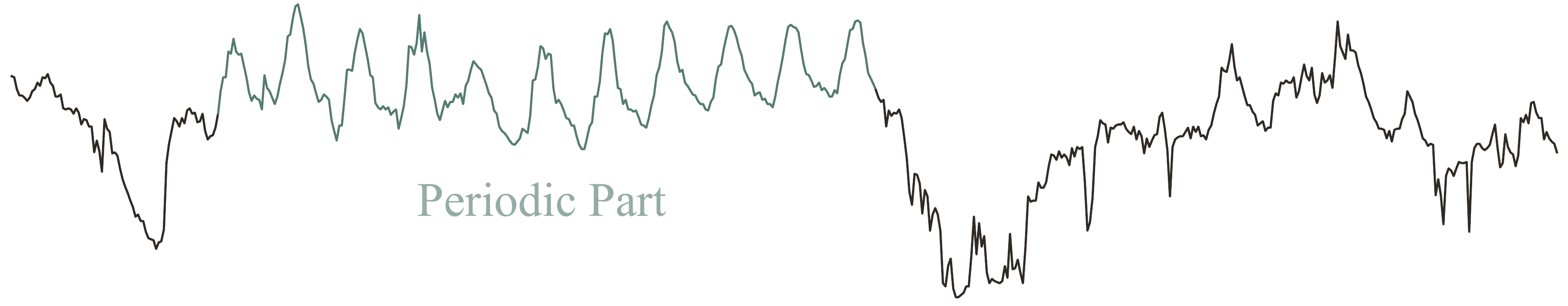}
}
\quad
\subfigure[Electricity  Variable: ‘3’]{
\includegraphics[width=1.5in]{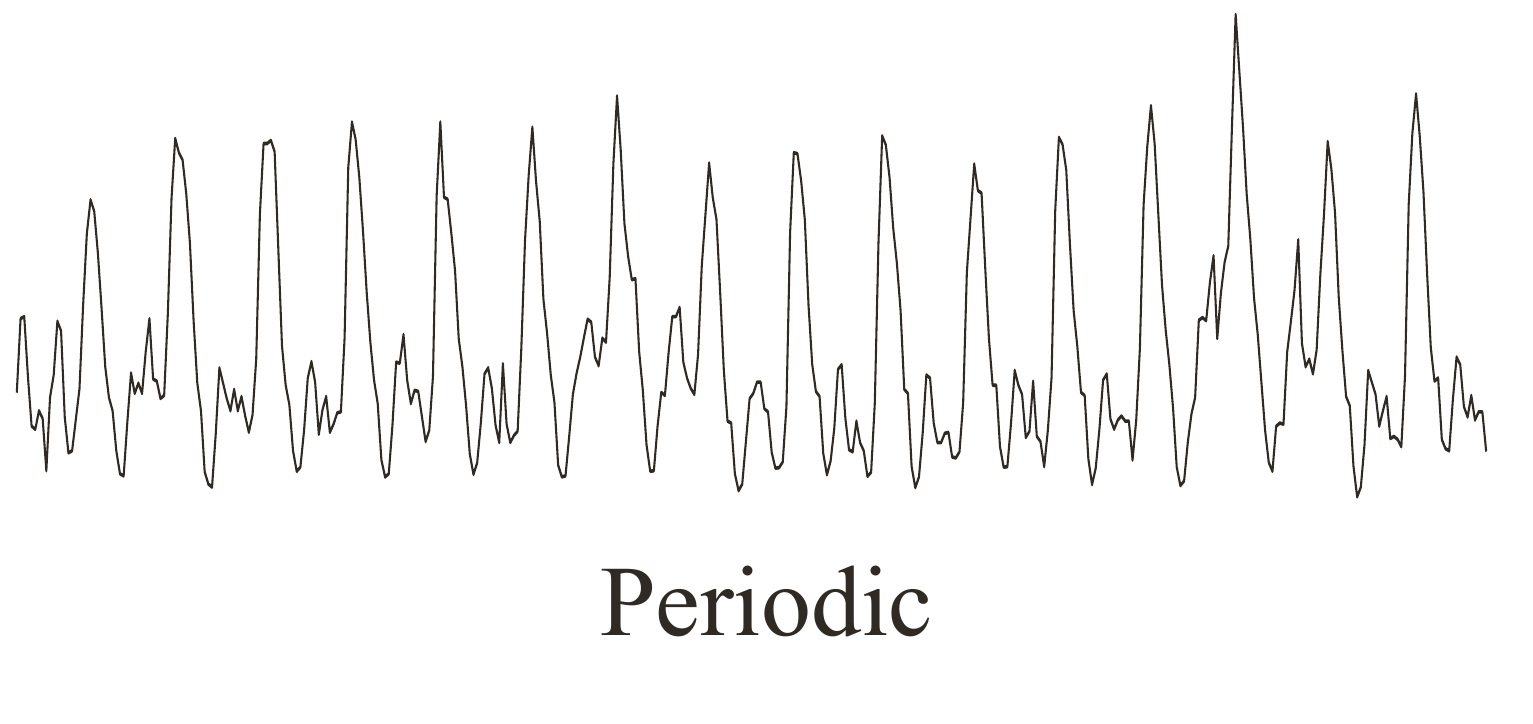}
}
\subfigure[Electricity  Variable: ‘125’]{
\includegraphics[width=1.5in]{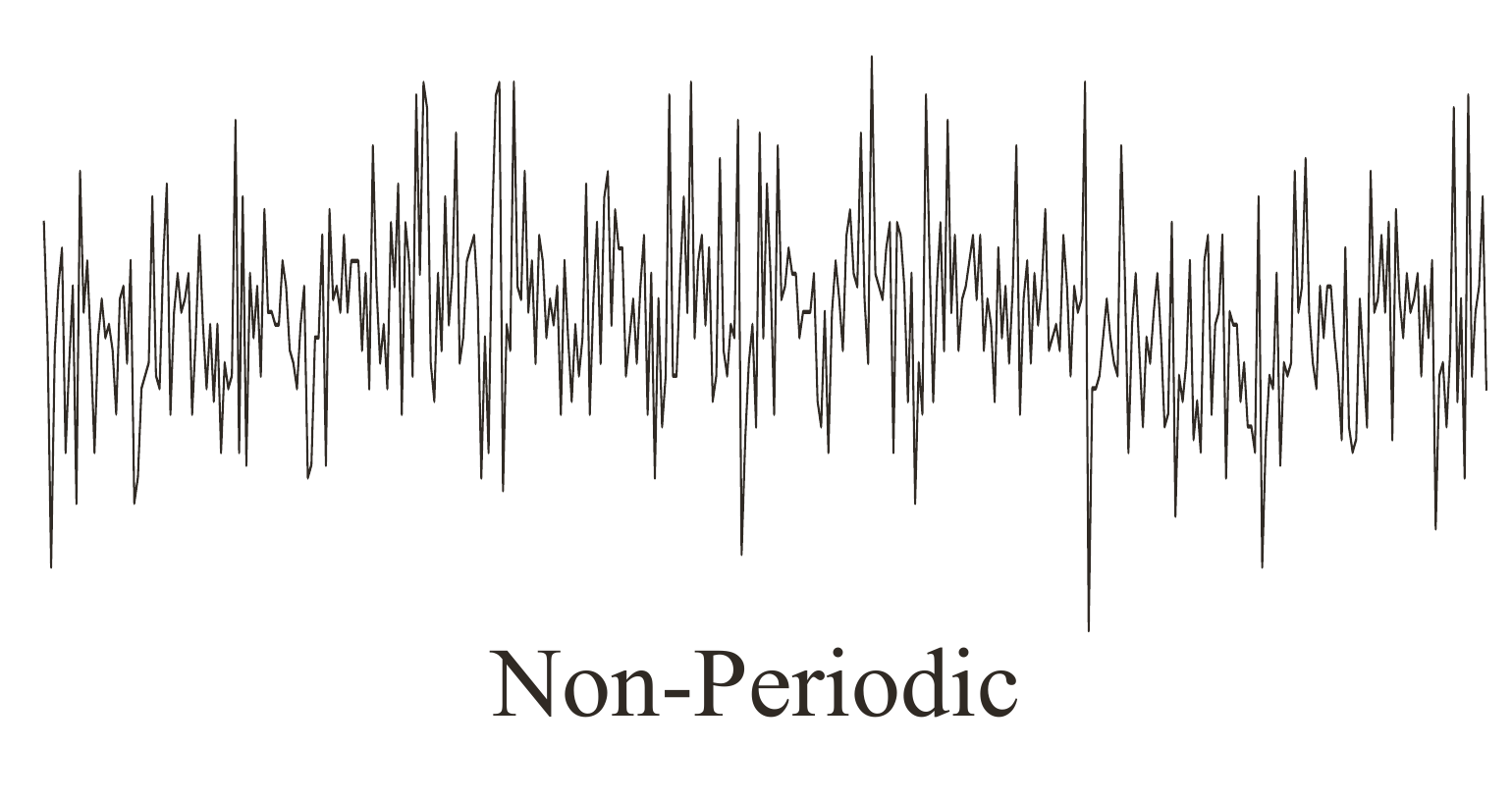}
}
\caption{Real-world time series with distinct periodic pattern.
}
\label{fig_intro}
\end{figure}

Real-world time series display differentiated periodic patterns. As illustrated in Figure \ref{fig_intro}, different segments of the same time series, as well as distinct time series within the same dataset, exhibit periodic characteristics with noticeable variations. Periodic time series typically exhibit more global dependencies, where the values of the series at one point in time are influenced by the values at other points in time that are separated by a specific interval. Non-periodic time series, on the other hand, exhibit more local dependencies. In this case, the values at one point in time may be influenced by nearby values in time, but there may not be a consistent and predictable relationship over longer periods. 

Time domain and frequency domain are two fundamental representations used to analyze time series data. In time domain, the analysis focuses on the variations in amplitude over time, allowing for the identification of local dependencies and transient behavior within the signal. Conversely, frequency domain analysis aims to represent time series in terms of their frequency components, providing insights into global dependencies and spectral characteristics of the data. Combining the advantages of both domains is a promising approach to address the challenge of dealing with the mixing of distinct periodic properties in real-world time series, but \textit{\textbf{how to effectively combine the advantages of time domain and frequency domain ($Q_1$)}} remains a challenge.

Early research predominantly focused on time-domain analysis. Over time, numerous intricate methodologies and sophisticated models have been developed to capture and exploit information in time domain, frequently relying on approaches like different neural network architectures~\cite{zhou2021informer, zhang2022crossformer}. Compared to the advancements in time domain, there are still many unexplored areas in frequency domain. In recent years, some works have employed frequency domain to better handle global dependencies of time series \citep{zhou2022fedformer, yi2023frequency, zhou2022film}. Most these works incorporate the frequency domain as an intermediary module instead of directly forecasting time series in frequency domain. Directly forecasting in frequency domain enables leveraging more spectral information to enhance the accuracy of time series predictions. FITS \cite{xu2023fits} directly forecasts spectrum of complete series with spectrum of input series as input in the frequency domain, and achieves promising results through lightweight linear layers, which has demonstrated the effectiveness and potential of this approach.

However, there are still some challenges associated with direct spectrum prediction in the frequency domain.
One of these challenges is the \textit{\textbf{potential misalignment of frequencies between the input spectrum and the complete spectrum ($Q_2$)}}, resulting from the use of the Discrete Fourier Transform (DFT). This misalignment makes it challenging to accurately represent the information of specific frequencies in the complete spectrum within the input spectrum, leading to prediction inaccuracies. 
Another challenge lies in \textit{\textbf{how to effectively extract information about frequency combinations ($Q_3$)}}. Extracting spectral features is a complex task, as the harmonic series appeared in groups within the spectrum contain a significant amount of information. Determining how to effectively capture and utilize this information about frequency combinations remains an open question.

To address the challenges mentioned above, we propose \model. It incorporates a time domain module and a frequency module to simultaneously handle local and global dependencies. Additionally, we introduce a novel weighting mechanism that dynamically allocates weights to the two modules. Our solutions for above challenges together with the contribution of this paper are listed as follows:

\textbf{\textit{$A_1$}}: We propose the \textbf{\textit{Dominant Harmonic Series Energy Weighting}} , which is able to generate appropriate weights for the time domain and frequency domain modules based on the level of periodicity exhibited by the input series. This allows us to leverage the advantages of both domains effectively when dealing with time series with distinct periodic patterns. We theoretically and empirically prove the effectiveness of out proposed method, which also provides with an effective way to measure the periodicity of a time series.

\textbf{\textit{$A_2$}}: Drawing inspiration from signal processing techniques employed to enhance measurement accuracy \citep{gough1994fast, santamaria2000comparative, luo2016interpolated}, we introduce \textbf{\textit{Extended DFT}} to align the discrete frequencies of the input spectrum with the complete spectrum, improving the accuracy of representing specific frequencies.

\textbf{\textit{$A_3$}}: We introduce the attention mechanism into frequency domain and propose \textbf{\textit{Complex-valued Spectrum Attention (CSA)}}. This approach allows us to capture information from various frequency combinations, which provides with an \textbf{{effective way of applying attention in addressing frequency domain representations}}.

Extensive expirements on $8$ real-world datasets demonstrates that our proposed \model\ achieves more promising  results compared with other state-of-the-art methods.

\section{Related Work} \label{sec:related_work}
\textbf{Time domain forecasting methods.} 
Early time series forecasting predominantly employed traditional statistical methods, such as ETS \cite{holt2004forecasting} and ARIMA \cite{box1968some}, which are are foundational yet often limited in capturing complex temporal dynamics. 
In recent years, the advent of deep learning has significantly advanced this field, particularly for long-term forecasting. RNN-based methods \citep{hochreiter1997long, chung2014empirical, lai2018modeling} and CNN-based methods \citep{bai2018empirical, liu2022scinet} have demonstrated enhanced capabilities in extracting intricate patterns from time series data.
Furthermore, Transformer-based methods \citep{zhou2021informer, kitaev2020reformer, zhang2022crossformer, nie2022time} can discover the temporal dependence between time points through the attention mechanism.
Additionally, MLP-based methods \citep{oreshkin2019n, challu2022nhits, zeng2023transformers} provide a straightforward yet effective approach to learning non-linear temporal dependencies of time series.


\begin{figure*}[!t]
\centering
\includegraphics[width=\textwidth]{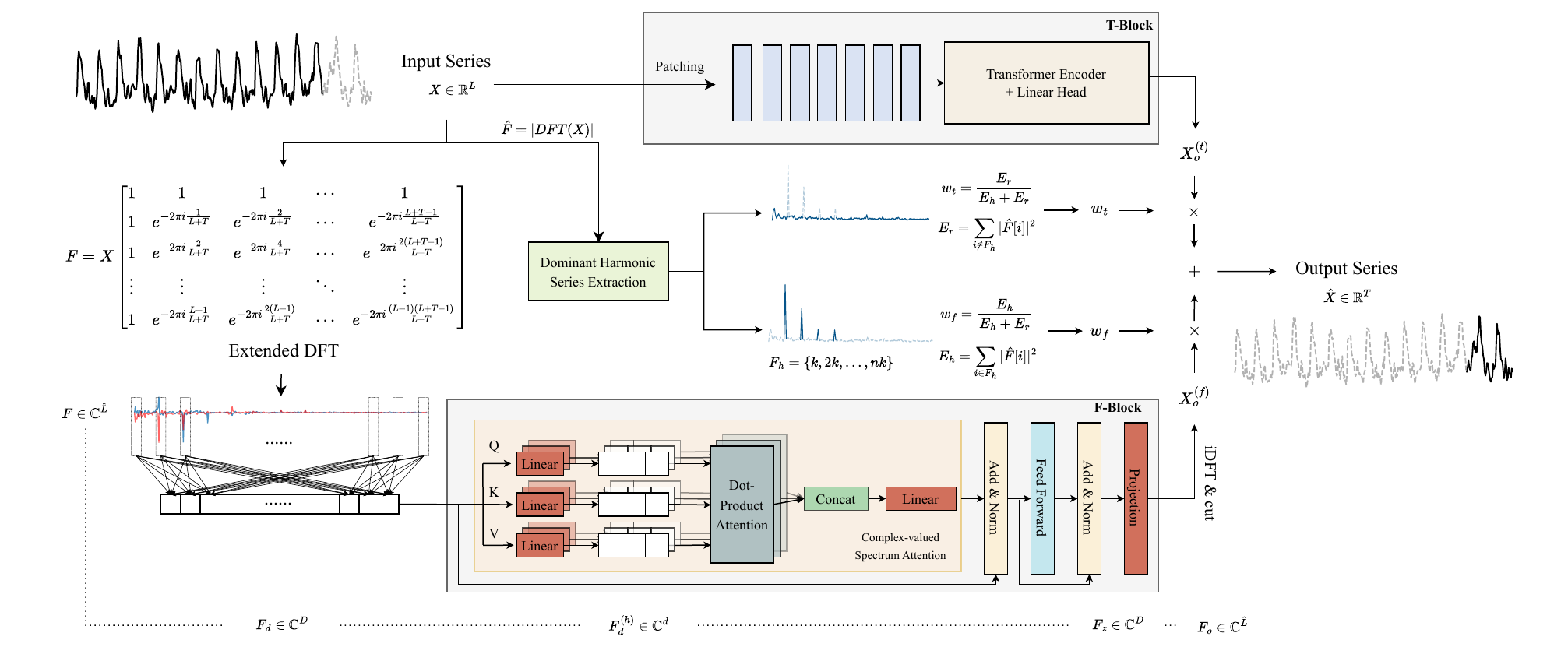}
\centering
\caption{ Model architecture of \model. \model\ is mainly composed of three sub-parts: 1) T-Block to capture local dependency from time domain; 2) F-Block to capture global dependency from frequency domain. The Extended DFT is used to generate frequency-aligned spectrum of input series. 3) The Dominant Harmonic Series Energy Weighting to allocate appropriate weights for F-Block and T-Block according to the periodic property of input series.
}
\label{fig_model}
\end{figure*}

\textbf{Frequency domain forecasting methods.}  
The frequency domain offers a unique perspective on time series data by revealing the underlying periodicity and frequency components. Within this domain, a variety of forecasting methods have emerged recently, specifically designed to harness these characteristics and effectively capture the global dependencies inherent in time series data.
Specifically, Autoformer \cite{wu2021autoformer} replaces the self-attention mechanism with an Auto-Correlation mechanism based on main periods detected using Fast Fourier Transform (FFT).
TimesNet \cite{wu2022timesnet} transforms 1D time series into 2D space according to the main periods detected from the topk frequency spectrum amplitude by FFT.
FilM \cite{zhou2022film} employs the Legendre Polynomials projection technique to approximate historical information, utilizes FFT for frequency feature extraction by Frequency Enhanced Layer.
FreDo \cite{sun2022fredo} proposes a simple yet strong model called AverageTile, transforming the tiled series to frequency domain with DFT to more accurately extract features from frequency domain.


\textbf{Complex number processing methods.}  
Extending time series analysis into the frequency domain necessitates transitioning from the real number field to the complex number field. The method of processing complex number is critical, as simplistic treatment as real numbers could result in missed analytical insights. StemGNN \cite{cao2020spectral} combines Graph Fourier Transform (GFT) for inter-series correlations and Discrete Fourier Transform (DFT) for temporal dependencies. However, its approach of transforming complex number into real numbers, and treating the real and imaginary parts separately, might not fully exploit the nuances of complex data interactions. In contrast, several methods have utilized complex-valued neural networks for more effective frequency domain transformations. FEDformer \cite{zhou2022fedformer} introduces a Frequency Enhanced Block, available in two versions based on DFT and Discrete Wavelet Transform (DWT), which transforms the original time series into the frequency domain using learnable complex-valued weights to better capture frequency domain relationships. FITS \cite{xu2023fits} employs FFT to convert time series to frequency domain and uses a complex-valued linear layer structure, supplemented with a low-pass filter to attenuate high-frequency noise. FreTS \cite{yi2023frequency} implements a specially designed real-valued MLP structure that mimics operations on complex number, managing both the real and imaginary parts of frequency coefficients simultaneously.

\section{Preliminary}
\subsection{Problem Formulation}
Long-term time series forecasting can be regarded as a sequence-to-sequence problem. Given a multi-variate time series $X=[x_1, x_2, \dots, x_L]\in \mathbb{R}^{N\times L}$, where $L$ is the look back window size and $N$ is the number of variates, the target is to predict the future $T$ steps $\hat{X}=[x_{L+1}, x_{L+2}, \dots, x_{L+T}]\in \mathbb{R}^{N\times T}$.

\subsection{Discrete Fourier Transform}
The Discrete Fourier Transform (DFT) is a mathematical transformation that converts a time series into its frequency domain representation, revealing the frequency components present in the data. 
Specifically, for an input series with length $L$ , the DFT spectrum is calculated as follows:
\begin{equation}
    F[k]=\sum_{n=0}^{L-1}X[n]e^{-2\pi i\frac{kn}{L}}, k=0,1,\dots,L-1.
    \label{DFT}
\end{equation}
Here, $F[k]$ represents the frequency-domain representation (the spectrum) at frequency index $k$, $X$ represents the time-domain input sequence, $L$ is the number of data points in the sequence, and $i$ is the imaginary unit ($i^2=-1$).

The Inverse Discrete Fourier Transform (iDFT) is the reverse operation of the DFT. It takes a frequency-domain signal, obtained through the DFT, and reconstructs the original time-domain signal:
\begin{equation}
    X[n]=\frac{1}{L}\sum_{k=0}^{L-1}F[k]e^{2\pi i\frac{kn}{L}}, n=0,1,\dots,L-1.
\end{equation}
\subsection{Energy in Time \& Frequency Domain}



The term `\textbf{Energy}' is commonly used in signal processing to refer to a concept that captures the overall strength or magnitude of a signal. In the context of a univariate time series represented by $x\in\mathbb{R}^L$, the energy of $x$ is defined as the sum of the squared values of its individual data points, expressed as $E_t=\sum_{n=0}^{L-1} x[n]^2$ \cite{Oppenheim2015Signal}. Denoting the Discrete Fourier Transform (DFT) spectrum of $x$ by $F\in\mathbb{C}^L$, we can define the energy of $F$ in a similar manner as $E_f=\sum_{n=0}^{L-1} |F[n]|^2$.

The energy distribution within the spectrum of a time series provides valuable insights into the presence and strength of periodic patterns. When a significant amount of energy is concentrated at dominant frequencies, it indicates the presence of strong periodic behavior in the data. On the other hand, a more uniform energy distribution across the spectrum suggests a lack of pronounced periodicity.

\section{Proposed Method}
 
 The structure of our proposed \model\ is depicted in Figure \ref{fig_model}. We employ a channel-independence scheme, which has been demonstrated to be effective by \citet{nie2022time}, to prevent the mixture of spectra from different channels. Mixing such spectra could have a detrimental impact since channels may possess distinct global patterns. 

For an input univariate series $X\in \mathbb{R}^{L}$, the T-Block directly processes it in the time domain, resulting in an output of $X_o^{(t)}\in \mathbb{R}^{T}$. We utilize the Extended Discrete Fourier Transform (DFT) to convert $X$ into the frequency domain, generating an extended spectrum denoted as $F\in\mathbb{C}^{L+T}$. After transforming the spectrum back to the time domain using the inverse DFT (iDFT), the F-Block produces the output $X_o^{(f)}\in \mathbb{R}^{T}$.
Finally, the output of the T-Block, $X_o^{(t)}$, and the output of the F-Block, $X_o^{(f)}$, are combined with weights to generate the final output $\hat{X}\in \mathbb{R}^{T}$. These weights are derived from the Dominant Harmonic Series Energy Weighting. The entire structure is formalized as:
\begin{equation}
    \begin{aligned}
        F &= \textbf{Extended DFT}(X),\\
        X_o^{f} &= \textbf{F-Block}(F),\\
        X_o^{t} &= \textbf{T-Block}(X),\\
        w_t, w_f &= \textbf{Weighting}(X),\\
        \hat{X} &= w_t X_o^{(t)} + w_f X_o^{(f)}.
    \end{aligned}
    \label{model_formula}
\end{equation}

In the following subsections, we will provide detailed explanations of the different sub-parts comprising \model.

\subsection{Extended DFT}
\label{Extended DFT}
\begin{figure}[htbp]
\centering
\includegraphics[width=3.3in]{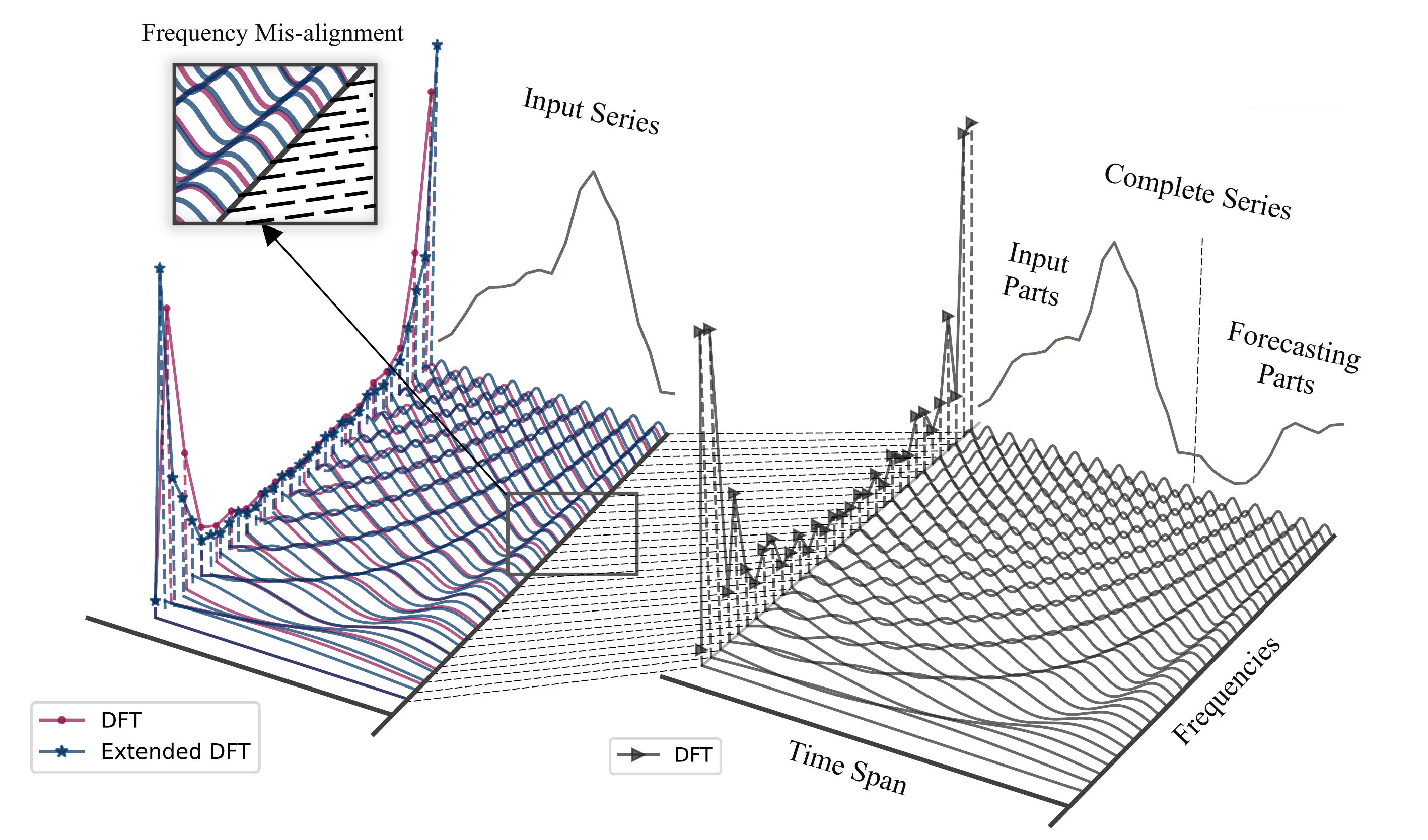}
\centering
\caption{By employing Extended DFT, it becomes possible to obtain the spectrum of the input series with a discrete frequency group that aligns with the DFT spectrum of the complete series.}
\label{fig_EDFT}
\end{figure}

 As illustrated in Section \ref{introduction}, using the traditional DFT may introduce a misalignment of frequencies between the spectra of input series and the entire series (as depicted in Figure \ref{fig_EDFT}). Consequently, forecasting models built on the input series may lack access to complete and accurate frequency information, leading to less accurate predictions for the complete series.

To address this issue, we propose Extended DFT, which overcomes the limitation imposed by the input length. This allows us to obtain an input spectrum that aligns with the DFT frequency group of the complete series.
Specifically, based on Eq.~\ref{DFT}, we replace the original complex exponential basis with the DFT basis of complete series:
\begin{equation} 
    F[k]=\sum_{n=0}^{L-1}x[n]e^{-2\pi i\frac{kn}{L+T}}, k=0,1,\dots,L+T-1.
    \label{eDFT}
\end{equation}



 By using Eq. \ref{eDFT}, we obtain a spectrum of length $L+T$ that aligns with the DFT spectrum of the complete series. For real-valued time series, the conjugate symmetry of the output spectrum is an important property of the DFT. Exploiting this property, we can reduce the computational cost by considering only the first half of the output spectrum, as the second half provides redundant information. Theorem \ref{theorem:1} demonstrates that Extended DFT maintains this property.

\begin{myTheo}  \label{theorem:1}
Let $F \in \mathbb{C}^{L+T}$ represent the output spectrum of the input series $X \in \mathbb{R}^L$ obtained through the Extended DFT, where $L$ denotes the input length and $T$ denotes the prediction length. We have the following statement holds
\begin{equation}
\begin{aligned}
\operatorname{Re}(F[k])&=\operatorname{Re}(F[L+T-k]),\\
\operatorname{Im}(F[k])&=-\operatorname{Im}(F[L+T-k]),  
\end{aligned}
\end{equation}
which implies that $F$ exhibits conjugate symmetry. Here, $k=[1,2,\dots,L+T-1]$, $\operatorname{Re}(\cdot)$ represents the real part, and $\operatorname{Im}(\cdot)$ represents the imaginary part.
\end{myTheo}
See Appendix \ref{proof_theorem1} for a detailed proof.


\subsection{F-Block} 
\label{F-Block}
The architecture of F-Block is based on the Transformer Encoder, with all parameters to be complex-valued. Notably, all calculations within the F-Block are performed in the complex number field. The F-Block takes a univariate spectrum $F\in\mathbb{C}^{\hat{L}}$ as input, which is generated by the Extended DFT. Here, $\hat{L}=\lfloor(L+T)/2\rfloor+1$ represents the length of the spectrum.


 Firstly, we utilize RevIN \cite{kim2021reversible} to process the input spectrum $F$. While RevIN was originally developed to address distribution shift in the time domain, we have found it to be effective in handling frequency domain spectra as well. This approach enables us to transform the spectra of series with distinct global features into a similar distribution. To accomplish this, we normalize $F$ by subtracting its mean and dividing it by its standard deviation. The standard deviation is calculated using $|F|\in\mathbb{R}^{\hat{L}}$. Afterwards, we add them back at the end of the structure. Next, we embed $F\in\mathbb{C}^{\hat{L}}$ into $F_d\in\mathbb{C}^D$ using a trainable linear projection $W_{emb}\in \mathbb{C}^{D\times\hat{L}}$. Since there are few chronological dependencies within the frequency domain spectrum, we disable the use of positional embeddings.

\textbf{Complex-valued Spectrum Attention}. We applied a modified multi-head attention mechanism rather than the original one in \citet{vaswani2017attention}. For each head $h=1,2,\dots,H$, it projects the embedded spectrum $F_d\in\mathbb{C}^{D}$ to dimension $d$ across the spectrum dimension with trainable projections.
Specifically, $Q_h=F_d^TW_h^Q$, $K_h=F_d^TW_h^K$, $V_h=F_d^TW_h^V$, where $W_h^Q, W_h^K, W_h^V\in \mathbb{C}^{D\times d}$ are learnable parameters.
On each head we perform complex-valued Dot-Product Attention:
\begin{equation}
    \begin{aligned}
        \textbf{Attention}(Q_h, K_h, V_h)&=\textbf{Softmax} (|Q_hK_h^T|)V_h.\\
    \end{aligned}
\end{equation}

 Then the final output of Complex-valued Spectrum Attention is calculated as follows:
\begin{equation}
    \begin{aligned}
        \text{head}_h &= \textbf{Attention}(Q_h, K_h, V_h),\\
        \text{CSA}(F_d)&=\textbf{Concat}(\text{head}_1, \dots, \text{head}_H)^TW_O,
    \end{aligned}
\end{equation}
where $W_O\in \mathbb{C}^{hd\times D}$ are learnable parameters.

We follow the design of LayerNorm and FeedForward layers with residual connections in Transformer, with expanding to complex number field. 
After $M$ layers of Encoder, the output $F_z$ is linearly projected to $F_o\in\mathbb{C}^{\hat{L}}$.
We convert $F_o$ to time domain using iDFT, taking the last $T$ points (the forecasting part) as the final output of F-Block $X_o^{(f)}\in\mathbb{R}^T$.

The F-Block applies a fully Complex-Valued Neural Network (CVNN) architecture, we further discuss the effectiveness of CVNN in Appendix \ref{cvnn}.

\subsection{T-Block} The T-Block is responsible for capturing local dependency in time series, which can be easier to be handled in time domain. Patching is an intuitive and effective way to capture local dependency in time series. Following \cite{nie2022time}, We employ patching together with the Transformer Encoder architecture as T-Block. It first devides the input series $X\in\mathbb{R}^L$ into patches $X_p\in\mathbb{R}^{P\times N}$, where $P$ is the length of patches and $N$ is the number of patches. The time series patches $X_p$ are then embed and fed into the Transformer Encoder. With a linear projection, it generates the final output $X_o^{(t)}\in\mathbb{R}^T$. We also employ RevIN in T-Block to deal with distribution shift problem.

\subsection{Dominant Harmonic Series Energy Weighting}
\label{dominant}

\begin{figure}[h]
\centering
\includegraphics[width=3in]{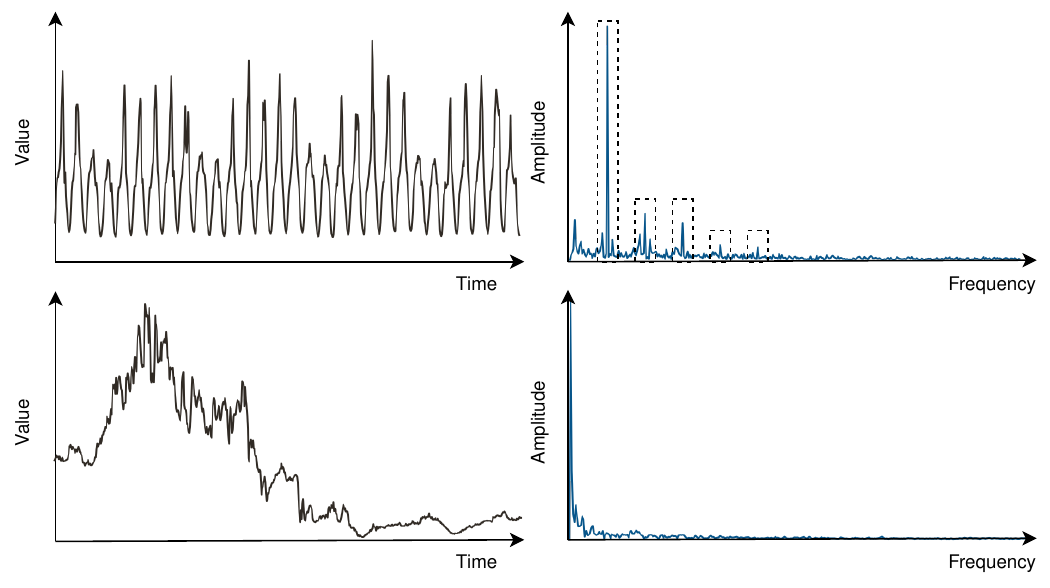}
\centering
\caption{\textbf{Top:} Periodic time series, sampled from Traffic dataset. It can be observed that there exists an obvious harmonic group. \textbf{Bottom:} Non-periodic time series, sampled from Weather dataset. There are no obvious harmonic group in non-periodic series.}
\label{fig_Harmonic}
\end{figure}

As depicted in Figure \ref{fig_Harmonic}, periodic time series consistently exhibits the existence of at least one harmonic group in their frequency domain spectrum, wherein the dominant harmonic group exhibits the highest concentration of spectrum energy. Conversely, this characteristic is seldom witnessed in the spectrum of non-periodic time series, where the distribution of energy is more uniform. The following Theorem \ref{theorem:2} shows that the extent of concentration of energy within the dominant harmonic series in the frequency spectrum can reflect the periodicity of a time series.

\begin{myTheo} \label{theorem:2}
Let $f(x)$ denote a continuous time series defined on the interval $[0,L]$. We denote the discrete form of length-$L$ realizations of $f(x)$, sampled at an interval of 1, as $X = [f(0), f(1), \dots, f(L-1)]$. $f(x)$ can be decomposed into two parts: $f(x)=f_p(x)+f_r(x)$, where $f_p(x)$ is a strictly periodic function with a period of $\tau$ (assuming $L=k\tau$ and $L, k, \tau\in \mathbb{N^*}$), i.e. $f_p(x-\tau)=f_p(x), \forall x\in[\tau, L]$. Let $E_p=\sum_{x=0}^{L-1}f_p^2(x) $ denotes the energy of periodic part and $E_r=\sum_{x=0}^{L-1}f_r^2(x)$ denotes the energy of the remaining part. 
Let $\lambda = E_p/E_r$, where $\lambda$ represents the magnitude of the periodic signal compared to the remaining signal, and varies with different dividing methods. 
Considering the frequency resolution limitations imposed by the sampling interval, we assume that the energy in the frequency component above the Nyquist frequency ($\pi$) is small enough to be neglected.
For simplicity, we additionally assume $\sum_{x=0}^{L-1} f(x)=0$. Under these assumptions, we have
\begin{equation} \label{eq:lowerbound}
    \frac{E_h}{E_f}=\frac{\sum_{n=0}^{T-1} |F[nk]|^2}{\sum_{n=0}^{L-1} |F[n]|^2 }\geq \frac{\lambda-2\sqrt{\lambda}}{\lambda - 2\sqrt{\lambda} + 1},
\end{equation}
where $E_h$ represents the total energy of the dominant harmonic group with a basis frequency $k=L/\tau$, $E_f$ represents the total energy of the whole spectrum, and $F(\cdot)$ represents the DFT components at a specific frequency.

\end{myTheo}
See Appendix \ref{proof_theorem2} for a detailed proof. The ratio $E_h/E_f$ in Eq. \ref{eq:lowerbound} serves as a metric for quantifying the concentration of energy within the dominant harmonic series in the spectrum of a time series. Theorem \ref{theorem:2} establishes a lower bound for this ratio, which increases monotonically with $\lambda$ for $\lambda>1$. Intuitively, when a time series exhibits a more prominent periodic pattern, it can be decomposed into components that correspond to a larger value of $\lambda$. Consequently, our theorem implies that such a time series possesses a higher concentration of energy within dominant harmonic series.


Based on this property, we aim to employ the energy proportion of the dominant harmonic series as a metric for quantifying the extent to which a time series exhibits periodicity.
To identify the dominant harmonic series, the crucial task is to determine the fundamental frequency. In addressing this challenge, we can draw insights from audio processing methodologies. Several methods can be employed for this purpose: 1) Naive method which identifies the frequency with the highest amplitude value as the fundamental frequency. 2) Rule-based pitch detection algorithms such as YIN\cite{de2002yin} and SWIPE\cite{camacho2008sawtooth}. 3) Data-driven pitch detection algorithms such as CREPE\cite{kim2018crepe} and SPICE\cite{gfeller2020spice}.


Specifically, we begin by identifying the fundamental frequency through one of the aforementioned methods. Next, We consider this component along with its $n$ harmonics, namely $F[2k], \dots, F[nk]$, and calculate the total energy $E_h$.
Subsequently, we determine the weights for the F-Block by computing the ratio $E_h/E_f$, where $E_f$ represents the total energy of the spectrum. The detailed procedure is presented in Algorithm \ref{algo:DHW}. 

\begin{algorithm}[htbp]  
   \caption{Dominant Harmonic Series Energy Weighting}
   \label{alg:1}
\begin{algorithmic}[1] \label{algo:DHW}
   \REQUIRE Time series $x$, length $L$, number of harmonics $N$
   \ENSURE The weight $w_f$, $w_t$ of the F-Block and T-Block
   \STATE $F=\text{DFT}(x)$ 
   \STATE $k=\text{Pitch Detection}(F)$ 
   \STATE $E_f=\sum_{i=0}^{\lfloor L/2\rfloor+1} |F[i]|^2$
   \STATE $E_h=\sum_{n=1}^{N} |F[nk]|^2$
   \STATE $w_f=E_h/E_f$ 
   \STATE $w_t=1-w_f$
\end{algorithmic}
\end{algorithm}
Through our extensive attempts, we have observed that the naive method demonstrates commendable accuracy across a majority of real-world time series datasets, all while maintaining a low computational cost. In contrast, alternative approaches are hindered by the issue of computational complexity. Additionally, data-driven methods necessitate labeled pitch data, which is often challenging to obtain, posing a significant obstacle to their practical utilization. In the experiments part, we adopt the naive method as default for detecting fundamental frequency in \model\ . Further exploration of employing various pitch detection algorithms is discussed in Appendix \ref{detection}.

\begin{table*}[htbp]
  \caption{Multi-variate long-term time series forecasting results on 8 datasets. The best results are in bold and the second best results are underlined. We only display the average results of all prediction lengths $T\in \{96, 192, 336, 720\}$ here, see Table \ref{full_multi}  for full results.}
  \label{multi}
  \vskip 0.05in
  \centering
  \begin{threeparttable}
  \begin{small}
  \renewcommand{\multirowsetup}{\centering}
  \setlength{\tabcolsep}{0.8pt}
  \begin{tabular}{c|c|cc|cc|cc|cc|cc|cc|cc|cc|cc|cc|cc|cc|cc}
    \toprule
\multicolumn{2}{c}{Models} &  
\multicolumn{2}{c}{\rotatebox{0}{\scalebox{0.8}{{\model}}}} &
\multicolumn{2}{c}{\rotatebox{0}{\scalebox{0.8}{{Autoformer}}}} &
\multicolumn{2}{c}{\rotatebox{0}{\scalebox{0.8}{{FEDformer}}}} &
\multicolumn{2}{c}{\rotatebox{0}{\scalebox{0.8}{{iTransformer}}}} &
\multicolumn{2}{c}{\rotatebox{0}{\scalebox{0.8}{{Crossformer}}}} &
\multicolumn{2}{c}{\rotatebox{0}{\scalebox{0.8}{{PatchTST}}}} &
\multicolumn{2}{c}{\rotatebox{0}{\scalebox{0.8}{{TimesNet}}}} &
\multicolumn{2}{c}{\rotatebox{0}{\scalebox{0.8}{{SCINet}}}} &
\multicolumn{2}{c}{\rotatebox{0}{\scalebox{0.8}{{DLinear}}}} &
\multicolumn{2}{c}{\rotatebox{0}{\scalebox{0.8}{{FiLM}}}} &
\multicolumn{2}{c}{\rotatebox{0}{\scalebox{0.8}{{FreTS}}}} &
\multicolumn{2}{c}{\rotatebox{0}{\scalebox{0.8}{{FITS}}}} &
\multicolumn{2}{c}{\rotatebox{0}{\scalebox{0.8}{{StemGNN}}}} \\
    
    \cmidrule(lr){3-4}\cmidrule(lr){5-6}\cmidrule(lr){7-8}\cmidrule(lr){9-10}\cmidrule(lr){11-12}\cmidrule(lr){13-14}\cmidrule(lr){15-16}
    \cmidrule(lr){17-18}\cmidrule(lr){19-20}\cmidrule(lr){21-22}\cmidrule(lr){23-24}\cmidrule(lr){25-26}\cmidrule(lr){27-28}
    \multicolumn{2}{c}{} & \scalebox{0.68}{MSE} & \scalebox{0.68}{MAE} & \scalebox{0.68}{MSE} & \scalebox{0.68}{MAE} & \scalebox{0.68}{MSE} &  \scalebox{0.68}{MAE} & \scalebox{0.68}{MSE} & \scalebox{0.68}{MAE} & \scalebox{0.68}{MSE} & \scalebox{0.68}{MAE} & \scalebox{0.68}{MSE} & \scalebox{0.68}{MAE} & \scalebox{0.68}{MSE} & \scalebox{0.68}{MAE} & \scalebox{0.68}{MSE} & \scalebox{0.68}{MAE} & \scalebox{0.68}{MSE} & \scalebox{0.68}{MAE} & \scalebox{0.68}{MSE} & \scalebox{0.68}{MAE} & \scalebox{0.68}{MSE} & \scalebox{0.68}{MAE} & \scalebox{0.68}{MSE} & \scalebox{0.68}{MAE} & \scalebox{0.68}{MSE} & \scalebox{0.68}{MAE}  \\ \midrule

\multicolumn{2}{c|}{ETTh1} 
&\boldres{\scalebox{0.78}{0.500}} &\boldres{\scalebox{0.78}{0.508}} &\scalebox{0.78}{0.672} &\scalebox{0.78}{0.608} &\scalebox{0.78}{0.649} &\scalebox{0.78}{0.601} &\scalebox{0.78}{0.538} &\scalebox{0.78}{0.528} &\scalebox{0.78}{0.529} &\scalebox{0.78}{0.533} &\secondres{\scalebox{0.78}{0.528}} &\secondres{\scalebox{0.78}{0.519}} &\scalebox{0.78}{0.616} &\scalebox{0.78}{0.577} &\scalebox{0.78}{0.837} &\scalebox{0.78}{0.674} &\scalebox{0.78}{0.533} &\scalebox{0.78}{0.532} &\scalebox{0.78}{0.596} &\scalebox{0.78}{0.549} &\scalebox{0.78}{0.550} &\scalebox{0.78}{0.533} &\scalebox{0.78}{1.641} &\scalebox{0.78}{0.905} &\scalebox{0.78}{0.703} &\scalebox{0.78}{0.639} \\
\midrule
\multicolumn{2}{c|}{ETTh2} 
&\boldres{\scalebox{0.78}{0.229}} &\boldres{\scalebox{0.78}{0.335}} &\scalebox{0.78}{0.265} &\scalebox{0.78}{0.361} &\scalebox{0.78}{0.262} &\scalebox{0.78}{0.359} &\scalebox{0.78}{0.249} &\scalebox{0.78}{0.344} &\scalebox{0.78}{0.248} &\scalebox{0.78}{0.355} &\scalebox{0.78}{0.241} &\secondres{\scalebox{0.78}{0.336}} &\scalebox{0.78}{0.277} &\scalebox{0.78}{0.364} &\scalebox{0.78}{0.321} &\scalebox{0.78}{0.403} &\scalebox{0.78}{0.238} &\scalebox{0.78}{0.344} &\scalebox{0.78}{0.267} &\scalebox{0.78}{0.352} &\secondres{\scalebox{0.78}{0.236}} &\scalebox{0.78}{0.341} &\scalebox{0.78}{0.413} &\scalebox{0.78}{0.452} &\scalebox{0.78}{0.393} &\scalebox{0.78}{0.456} \\
\midrule
\multicolumn{2}{c|}{ETTm1} 
&\boldres{\scalebox{0.78}{0.388}} &\boldres{\scalebox{0.78}{0.425}} &\scalebox{0.78}{0.809} &\scalebox{0.78}{0.633} &\scalebox{0.78}{0.539} &\scalebox{0.78}{0.540} &\scalebox{0.78}{0.423} &\scalebox{0.78}{0.442} &\scalebox{0.78}{0.405} &\scalebox{0.78}{0.440} &\secondres{\scalebox{0.78}{0.396}} &\scalebox{0.78}{0.429} &\scalebox{0.78}{0.457} &\scalebox{0.78}{0.473} &\scalebox{0.78}{0.475} &\scalebox{0.78}{0.473} &\scalebox{0.78}{0.413} &\secondres{\scalebox{0.78}{0.427}} &\scalebox{0.78}{0.457} &\scalebox{0.78}{0.449} &\scalebox{0.78}{0.417} &\scalebox{0.78}{0.437} &\scalebox{0.78}{0.617} &\scalebox{0.78}{0.549} &\scalebox{0.78}{0.644} &\scalebox{0.78}{0.599} \\
\midrule
\multicolumn{2}{c|}{ETTm2} 
&\secondres{\scalebox{0.78}{0.162}} &\boldres{\scalebox{0.78}{0.270}} &\scalebox{0.78}{0.185} &\scalebox{0.78}{0.296} &\scalebox{0.78}{0.175} &\scalebox{0.78}{0.283} &\scalebox{0.78}{0.171} &\scalebox{0.78}{0.279} &\scalebox{0.78}{0.171} &\scalebox{0.78}{0.288} &\scalebox{0.78}{0.165} &\scalebox{0.78}{0.273} &\scalebox{0.78}{0.176} &\scalebox{0.78}{0.283} &\scalebox{0.78}{0.180} &\scalebox{0.78}{0.290} &\boldres{\scalebox{0.78}{0.161}} &\secondres{\scalebox{0.78}{0.272}} &\scalebox{0.78}{0.173} &\scalebox{0.78}{0.278} &\scalebox{0.78}{0.167} &\scalebox{0.78}{0.279} &\scalebox{0.78}{0.197} &\scalebox{0.78}{0.305} &\scalebox{0.78}{0.239} &\scalebox{0.78}{0.357} \\
\midrule
\multicolumn{2}{c|}{Traffic} 
&\boldres{\scalebox{0.78}{0.403}} &\boldres{\scalebox{0.78}{0.279}} &\scalebox{0.78}{0.630} &\scalebox{0.78}{0.387} &\scalebox{0.78}{0.608} &\scalebox{0.78}{0.374} &\secondres{\scalebox{0.78}{0.420}} &\scalebox{0.78}{0.296} &\scalebox{0.78}{0.948} &\scalebox{0.78}{0.563} &\scalebox{0.78}{0.423} &\secondres{\scalebox{0.78}{0.295}} &\scalebox{0.78}{0.816} &\scalebox{0.78}{0.497} &\scalebox{0.78}{0.533} &\scalebox{0.78}{0.387} &\scalebox{0.78}{0.456} &\scalebox{0.78}{0.320} &\scalebox{0.78}{0.491} &\scalebox{0.78}{0.345} &\scalebox{0.78}{0.466} &\scalebox{0.78}{0.318} &\scalebox{0.78}{0.870} &\scalebox{0.78}{0.566} &\scalebox{0.78}{0.847} &\scalebox{0.78}{0.478} \\
\midrule
\multicolumn{2}{c|}{Electricity} 
&\boldres{\scalebox{0.78}{0.161}} &\boldres{\scalebox{0.78}{0.257}} &\scalebox{0.78}{0.215} &\scalebox{0.78}{0.327} &\scalebox{0.78}{0.222} &\scalebox{0.78}{0.335} &\secondres{\scalebox{0.78}{0.163}} &\secondres{\scalebox{0.78}{0.259}} &\scalebox{0.78}{0.188} &\scalebox{0.78}{0.285} &\scalebox{0.78}{0.170} &\scalebox{0.78}{0.270} &\scalebox{0.78}{0.288} &\scalebox{0.78}{0.375} &\scalebox{0.78}{0.184} &\scalebox{0.78}{0.289} &\scalebox{0.78}{0.173} &\scalebox{0.78}{0.275} &\scalebox{0.78}{0.185} &\scalebox{0.78}{0.284} &\scalebox{0.78}{0.173} &\scalebox{0.78}{0.273} &\scalebox{0.78}{0.313} &\scalebox{0.78}{0.409} &\scalebox{0.78}{0.230} &\scalebox{0.78}{0.334} \\
\midrule
\multicolumn{2}{c|}{Weather} 
&\boldres{\scalebox{0.78}{0.228}} &\secondres{\scalebox{0.78}{0.268}} &\scalebox{0.78}{0.328} &\scalebox{0.78}{0.371} &\scalebox{0.78}{0.321} &\scalebox{0.78}{0.370} &\scalebox{0.78}{0.238} &\scalebox{0.78}{0.272} &\scalebox{0.78}{0.230} &\scalebox{0.78}{0.288} &\secondres{\scalebox{0.78}{0.229}} &\boldres{\scalebox{0.78}{0.265}} &\scalebox{0.78}{0.265} &\scalebox{0.78}{0.296} &\scalebox{0.78}{0.267} &\scalebox{0.78}{0.299} &\scalebox{0.78}{0.247} &\scalebox{0.78}{0.300} &\scalebox{0.78}{0.270} &\scalebox{0.78}{0.301} &\scalebox{0.78}{0.248} &\scalebox{0.78}{0.298} &\scalebox{0.78}{0.318} &\scalebox{0.78}{0.340} &\scalebox{0.78}{0.320} &\scalebox{0.78}{0.390} \\
\midrule
\multicolumn{2}{c|}{Solar} 
&\secondres{\scalebox{0.78}{0.198}} &\secondres{\scalebox{0.78}{0.261}} &\scalebox{0.78}{0.747} &\scalebox{0.78}{0.649} &\scalebox{0.78}{0.285} &\scalebox{0.78}{0.392} &\scalebox{0.78}{0.219} &\scalebox{0.78}{0.272} &\boldres{\scalebox{0.78}{0.193}} &\boldres{\scalebox{0.78}{0.249}} &\scalebox{0.78}{0.205} &\scalebox{0.78}{0.272} &\scalebox{0.78}{0.325} &\scalebox{0.78}{0.362} &\scalebox{0.78}{0.264} &\scalebox{0.78}{0.327} &\scalebox{0.78}{0.258} &\scalebox{0.78}{0.321} &\scalebox{0.78}{0.296} &\scalebox{0.78}{0.328} &\scalebox{0.78}{0.231} &\scalebox{0.78}{0.290} &\scalebox{0.78}{0.392} &\scalebox{0.78}{0.450} &\scalebox{0.78}{0.235} &\scalebox{0.78}{0.301} \\

    \bottomrule
  \end{tabular}
    \end{small}
  \end{threeparttable}
\end{table*}

\section{Experiments}
We evaluate the performance of \model\ architecture through extensive experiments on 8 real-world datasets, including energy, weather, and traffic. We choose some state-of-the-art long-term time series forecasting models as baselines, including Autoformer~\citep{wu2021autoformer}, FEDformer~\citep{zhou2022fedformer}, iTransformer~\citep{liu2023itransformer}, Crossformer~\citep{zhang2022crossformer}, PatchTST~\citep{nie2022time}, TimesNet~\citep{wu2022timesnet}, SCINet~\citep{liu2022scinet}, DLinear~\citep{zeng2023transformers}, FiLM~\citep{zhou2022film}, FreTS~\citep{yi2023frequency}, FITS~\citep{xu2023fits}, and StemGNN~\citep{cao2020spectral}. 
The codes are uploaded to \href{https://github.com/YHYHYHYHYHY/ATFNet}{https://github.com/YHYHYHYHYHY/ATFNet}. to achieve reproducibility.

For all the models, we test the performance under 4 prediction length $T\in\{96, 192, 336, 720\}$. As for the look-back window size $L$, we run three distinct sizes $L\in\{96, 192, 336\}$ and always choose the best results to deal with the problem that not all models are suitable for the same look-back window size.

\begin{figure}[t]
\centering
\subfigure[Traffic]{
\includegraphics[width=\linewidth]{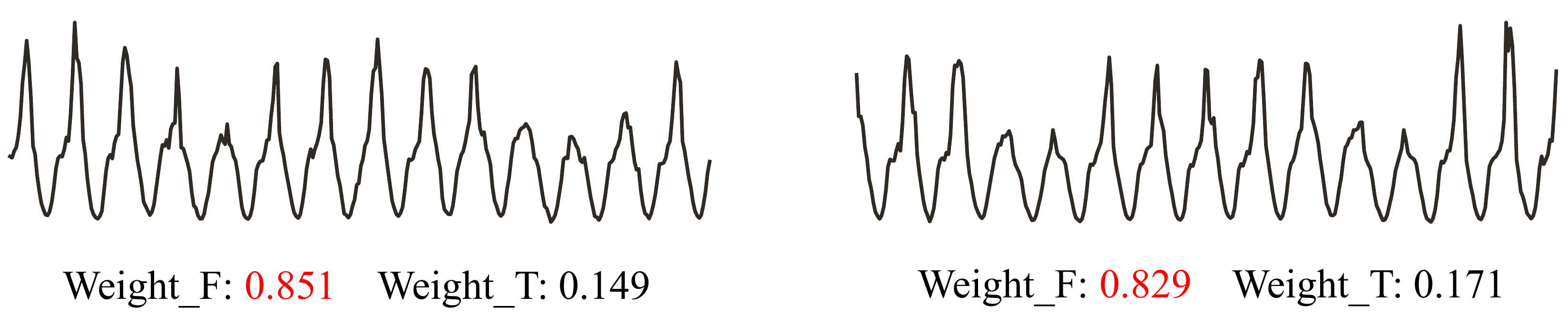}
}
\quad
\subfigure[Electricity]{
\includegraphics[width=\linewidth]{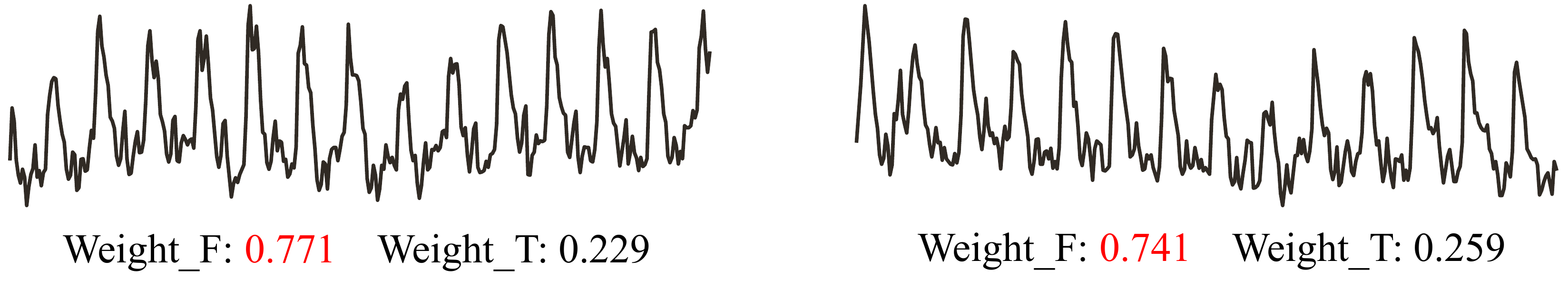}
}
\quad
\subfigure[Weather]{
\includegraphics[width=\linewidth]{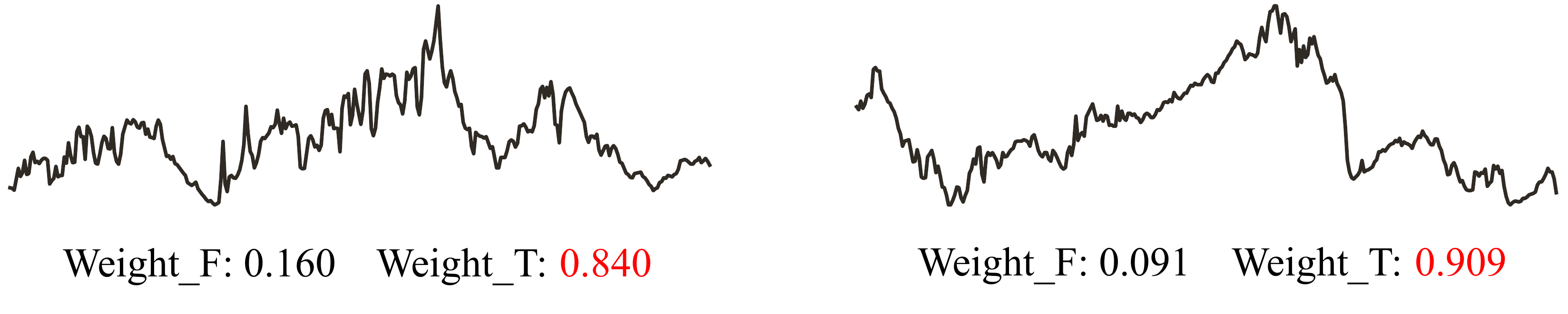}
}
\quad
\subfigure[ETTh1]{
\includegraphics[width=\linewidth]{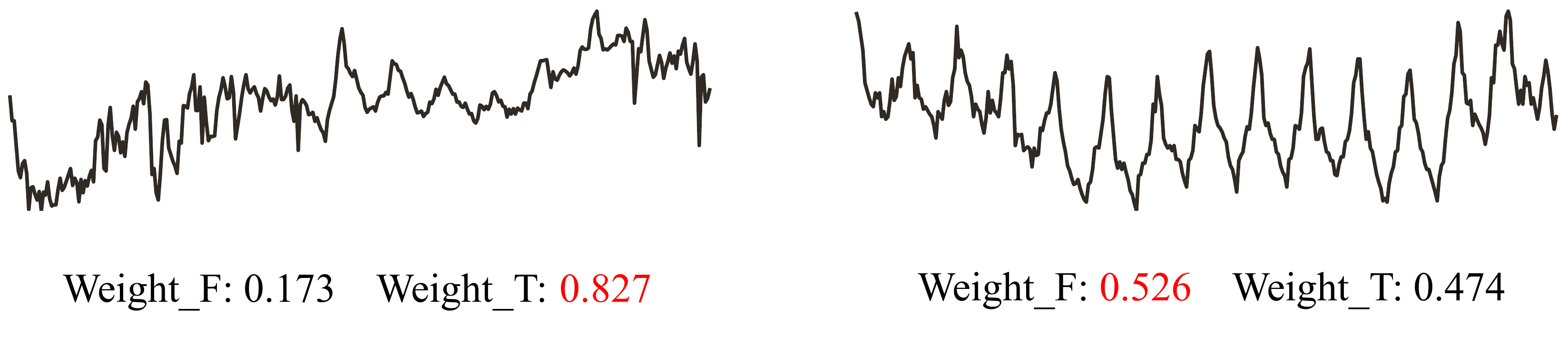}
}
\caption{Examples of weights allocation for distinct time series from 4 different datasets.
}
\label{exp_har}
\end{figure}

\subsection{Long-term Time Series Forecasting}



Comprehensive forecasting results for multi-variate and uni-variate forecasting task are listed in Table~\ref{multi} and Table~\ref{uni} with the best in red and the second underlined. The lower MSE/MAE indicates the more accurate prediction result. We adopt the naive method to identify the fundamental frequency for \model. 

For the multi-variate long-term forecasting task, \model\ achieves the most promising results on both MSE and MAE compared to other baseline models. Through the full results in Table \ref{full_multi}, it can be further observed that when the forecasting horizon is large (720), \model\ achieves the best results. Additionally, even in the remaining dataset (ETTm2), \model\ achieves the second-best result. These results highlight the effectiveness of \model\ in handling long-term time series forecasting and capturing the complex relationships and dependencies that arise over extended time periods. 

As for the addressing of datasets exhibiting distinct periodic patterns, we select 3 representive datasets for clarification.
\textbf{1) Traffic dataset}: This dataset exhibits periodic characteristics. As shown in Table \ref{full_multi}, \model\ consistently achieves the best results in terms of MSE and MAE across all settings, including four distinct forecasting horizons. Notably, \model\ exhibits a relative improvement of 4.0\% in  average MSE compared to the second best (also the current SOTA) iTransformer. The results in Table \ref{tab:ablation} also indicate that the F-Block, which leverages our proposed CSA to capture global dependencies in periodic time series, contributes significantly to the performance improvements.
\textbf{2) Weather dataset}: This dataset represents non-periodic time series. We acknowledge that \model\ does not achieve significantly promising results in the Weather dataset, as our proposed model does not include specific modifications tailored for handling non-periodic time series with few global dependencies. However, the design of including Transformer with patching (the idea borrowed from PatchTST) contributes a lot to remain good performance by effectively capturing important local patterns and relationships within the time series.
\textbf{3) ETTh1 dataset}: This dataset represents a mixture of both periodic and non-periodic time series. In this case, \model\ achieves relatively more advanced results, demonstrating a relative improvement of 5.3\% in average MSE compared to the second-best model, PatchTST. Table \ref{tab:ablation} also reveals that both the T-Block and F-Block individually do not achieve such results, indicating that our proposed ensemble method, the Dominant Harmonic Group Energy Weighting, effectively utilizes the advantages of both modules and contributes to a significant improvement when addressing datasets with a mixture of periodic and non-periodic time series.

We also experiment with univariate datasets where the results are provided in Table~\ref{uni}, which also demonstrates state-of-the-art performance.

\begin{table*}[htbp]
  \caption{Uni-variate long-term time series forecasting results on ETT datasets. The ETT datasets have a target feature `Oil Temperature', and we take it as the uni-variate time series for forecasting. The best results are in bold and the second best results are underlined. We only display the average results of all prediction lengths $T\in\{96, 192, 336, 720\}$ here, see Table \ref{full_uni} in Appendix for full results.}
  \label{uni}
  \vskip 0.05in
  \centering
  \begin{threeparttable}
  \begin{small}
  \renewcommand{\multirowsetup}{\centering}
  \setlength{\tabcolsep}{0.8pt}
  \begin{tabular}{c|c|cc|cc|cc|cc|cc|cc|cc|cc|cc|cc|cc}
    \toprule
\multicolumn{2}{c}{Models} &  
\multicolumn{2}{c}{\rotatebox{0}{\scalebox{0.8}{{\model}}}} &
\multicolumn{2}{c}{\rotatebox{0}{\scalebox{0.8}{{Autoformer}}}} &
\multicolumn{2}{c}{\rotatebox{0}{\scalebox{0.8}{{FEDformer}}}} &
\multicolumn{2}{c}{\rotatebox{0}{\scalebox{0.8}{{PatchTST}}}} &
\multicolumn{2}{c}{\rotatebox{0}{\scalebox{0.8}{{TimesNet}}}} &
\multicolumn{2}{c}{\rotatebox{0}{\scalebox{0.8}{{SCINet}}}} &
\multicolumn{2}{c}{\rotatebox{0}{\scalebox{0.8}{{DLinear}}}} &
\multicolumn{2}{c}{\rotatebox{0}{\scalebox{0.8}{{FiLM}}}} &
\multicolumn{2}{c}{\rotatebox{0}{\scalebox{0.8}{{FreTS}}}} &
\multicolumn{2}{c}{\rotatebox{0}{\scalebox{0.8}{{FITS}}}} &
\multicolumn{2}{c}{\rotatebox{0}{\scalebox{0.8}{{TFDNet}}}} \\
    
    \cmidrule(lr){3-4}\cmidrule(lr){5-6}\cmidrule(lr){7-8}\cmidrule(lr){9-10}\cmidrule(lr){11-12}\cmidrule(lr){13-14}\cmidrule(lr){15-16}
    \cmidrule(lr){17-18}\cmidrule(lr){19-20}\cmidrule(lr){21-22}\cmidrule(lr){23-24}
    \multicolumn{2}{c}{} & \scalebox{0.68}{MSE} & \scalebox{0.68}{MAE} & \scalebox{0.68}{MSE} & \scalebox{0.68}{MAE} & \scalebox{0.68}{MSE} &  \scalebox{0.68}{MAE} & \scalebox{0.68}{MSE} & \scalebox{0.68}{MAE} & \scalebox{0.68}{MSE} & \scalebox{0.68}{MAE} & \scalebox{0.68}{MSE} & \scalebox{0.68}{MAE} & \scalebox{0.68}{MSE} & \scalebox{0.68}{MAE} & \scalebox{0.68}{MSE} & \scalebox{0.68}{MAE} & \scalebox{0.68}{MSE} & \scalebox{0.68}{MAE} & \scalebox{0.68}{MSE} & \scalebox{0.68}{MAE} & \scalebox{0.68}{MSE} & \scalebox{0.68}{MAE}  \\ \midrule

\multicolumn{2}{c|}{ETTh1} 
&\scalebox{0.78}{0.118} &\scalebox{0.78}{0.270} &\scalebox{0.78}{0.124} &\scalebox{0.78}{0.279} &\scalebox{0.78}{0.129} &\scalebox{0.78}{0.284} &\scalebox{0.78}{0.134} &\scalebox{0.78}{0.285} &\scalebox{0.78}{0.126} &\scalebox{0.78}{0.277} &\scalebox{0.78}{0.158} &\scalebox{0.78}{0.313} &\secondres{\scalebox{0.78}{0.115}} &\secondres{\scalebox{0.78}{0.269}} &\scalebox{0.78}{0.124} &\scalebox{0.78}{0.279} &\boldres{\scalebox{0.78}{0.114}} &\boldres{\scalebox{0.78}{0.265}} &\scalebox{0.78}{0.212} &\scalebox{0.78}{0.361} &\scalebox{0.78}{0.138} &\scalebox{0.78}{0.293} \\
\midrule
\multicolumn{2}{c|}{ETTh2} 
&\boldres{\scalebox{0.78}{0.293}} &\boldres{\scalebox{0.78}{0.428}} &\secondres{\scalebox{0.78}{0.296}} &\secondres{\scalebox{0.78}{0.436}} &\scalebox{0.78}{0.315} &\scalebox{0.78}{0.445} &\scalebox{0.78}{0.332} &\scalebox{0.78}{0.452} &\scalebox{0.78}{0.317} &\scalebox{0.78}{0.447} &\scalebox{0.78}{0.429} &\scalebox{0.78}{0.519} &\scalebox{0.78}{0.302} &\scalebox{0.78}{0.440} &\scalebox{0.78}{0.305} &\scalebox{0.78}{0.438} &\scalebox{0.78}{0.346} &\scalebox{0.78}{0.460} &\scalebox{0.78}{0.690} &\scalebox{0.78}{0.650} &\scalebox{0.78}{0.386} &\scalebox{0.78}{0.491} \\
\midrule
\multicolumn{2}{c|}{ETTm1} 
&\boldres{\scalebox{0.78}{0.088}} &\boldres{\scalebox{0.78}{0.217}} &\scalebox{0.78}{0.112} &\scalebox{0.78}{0.258} &\scalebox{0.78}{0.105} &\scalebox{0.78}{0.244} &\scalebox{0.78}{0.089} &\secondres{\scalebox{0.78}{0.221}} &\scalebox{0.78}{0.090} &\scalebox{0.78}{0.222} &\scalebox{0.78}{0.094} &\scalebox{0.78}{0.230} &\secondres{\scalebox{0.78}{0.088}} &\scalebox{0.78}{0.221} &\scalebox{0.78}{0.098} &\scalebox{0.78}{0.234} &\scalebox{0.78}{0.091} &\scalebox{0.78}{0.226} &\scalebox{0.78}{0.120} &\scalebox{0.78}{0.264} &\scalebox{0.78}{0.098} &\scalebox{0.78}{0.232} \\
\midrule
\multicolumn{2}{c|}{ETTm2} 
&\boldres{\scalebox{0.78}{0.203}} &\boldres{\scalebox{0.78}{0.331}} &\scalebox{0.78}{0.298} &\scalebox{0.78}{0.417} &\scalebox{0.78}{0.243} &\scalebox{0.78}{0.369} &\secondres{\scalebox{0.78}{0.214}} &\secondres{\scalebox{0.78}{0.338}} &\scalebox{0.78}{0.229} &\scalebox{0.78}{0.354} &\scalebox{0.78}{0.236} &\scalebox{0.78}{0.363} &\scalebox{0.78}{0.221} &\scalebox{0.78}{0.350} &\scalebox{0.78}{0.259} &\scalebox{0.78}{0.377} &\scalebox{0.78}{0.225} &\scalebox{0.78}{0.348} &\scalebox{0.78}{0.357} &\scalebox{0.78}{0.458} &\scalebox{0.78}{0.252} &\scalebox{0.78}{0.368} \\

    \bottomrule
  \end{tabular}
    \end{small}
  \end{threeparttable}
\end{table*}

\subsection{Dominant Harmonic Series Energy Weighting}

The purpose of Dominant Harmonic Series Energy Weighting is to adjust weights allocating to F-Block and T-Block according to the periodic property of a certain time series. As depicted in Figure \ref{exp_har}, we take 8 time series from 4 different datasets to test whether our proposed Dominant Harmonic Series Energy Weighting can generate appropriate weights for distinct time series. We can observe that for time series with good periodic pattern, such as the time series in Traffic and Electricity dataset, are allocated more weights for F-Block. While for other time series without obvious periodic pattern, such as the time series in Weather dataset and the first time series in ETTh1 dataset, T-Block is allocated more weights.

\subsection{Ablation Study}
In this part, we conduct several experiments to analyze the contribution of different sub-modules of \model, and the results are listed in Table \ref{tab:ablation}. We designed 3 distinct versions of \model. 1) \model\ v1 uses traditional DFT instead of Extended DFT . 2) \model\ v2 use the simple average instead of Dominant Harmonic Series Energy Weighting. 3) the complex-valued network of \model\ is implemented with 2 independent real-valued channels in \model\ v3. We also include single T-Block and F-Block to test how \model\ utilize the advantages of both to enhancing forecasting performance.

\begin{table}[h]
  \caption{Ablation study results. The best results are in bold and the second best results are underlined. We only display the average results of all prediction lengths $T\in \{96, 192, 336, 720\}$ here, see Table \ref{full_ablation} in Appendix for full results.}
  \label{tab:ablation}
  \vskip 0.05in
  \centering
  \begin{threeparttable}
  \begin{small}
  \renewcommand{\multirowsetup}{\centering}
  \setlength{\tabcolsep}{0.8pt}
  \begin{tabular}{c|c|cc|cc|cc|cc|cc|cc}
    \toprule
\multicolumn{2}{c}{Models} &  
\multicolumn{2}{c}{\rotatebox{0}{\scalebox{0.8}{{ATFNet}}}} &
\multicolumn{2}{c}{\rotatebox{0}{\scalebox{0.8}{{F-Block}}}} &
\multicolumn{2}{c}{\rotatebox{0}{\scalebox{0.8}{{T-Block}}}} &
\multicolumn{2}{c}{\rotatebox{0}{\scalebox{0.8}{{ATFNet v1}}}} &
\multicolumn{2}{c}{\rotatebox{0}{\scalebox{0.8}{{ATFNet v2}}}} &
\multicolumn{2}{c}{\rotatebox{0}{\scalebox{0.8}{{ATFNet v3}}}}  \\
    
    \cmidrule(lr){3-4}\cmidrule(lr){5-6}\cmidrule(lr){7-8}\cmidrule(lr){9-10}\cmidrule(lr){11-12}\cmidrule(lr){13-14}
    \multicolumn{2}{c}{} & \scalebox{0.68}{MSE} & \scalebox{0.68}{MAE} & \scalebox{0.68}{MSE} & \scalebox{0.68}{MAE} & \scalebox{0.68}{MSE} &  \scalebox{0.68}{MAE} & \scalebox{0.68}{MSE} & \scalebox{0.68}{MAE} & \scalebox{0.68}{MSE} & \scalebox{0.68}{MAE} & \scalebox{0.68}{MSE} & \scalebox{0.68}{MAE}   \\ \midrule

\multicolumn{2}{c|}{ETTh1} 
&\boldres{\scalebox{0.78}{0.500}} &\boldres{\scalebox{0.78}{0.508}} &\scalebox{0.78}{0.526} &\scalebox{0.78}{0.521} &\scalebox{0.78}{0.528} &\scalebox{0.78}{0.519} &\secondres{\scalebox{0.78}{0.506}} &\secondres{\scalebox{0.78}{0.509}} &\scalebox{0.78}{0.511} &\scalebox{0.78}{0.511} &\scalebox{0.78}{0.510} &\scalebox{0.78}{0.512} \\
\midrule
\multicolumn{2}{c|}{ETTm1} 
&\boldres{\scalebox{0.78}{0.388}} &\boldres{\scalebox{0.78}{0.425}} &\scalebox{0.78}{0.408} &\scalebox{0.78}{0.433} &\scalebox{0.78}{0.396} &\scalebox{0.78}{0.429} &\scalebox{0.78}{0.393} &\scalebox{0.78}{0.428} &\scalebox{0.78}{0.393} &\scalebox{0.78}{0.427} &\secondres{\scalebox{0.78}{0.392}} &\secondres{\scalebox{0.78}{0.427}} \\
\midrule
\multicolumn{2}{c|}{Traffic} 
&\boldres{\scalebox{0.78}{0.403}} &\boldres{\scalebox{0.78}{0.279}} &\scalebox{0.78}{0.406} &\scalebox{0.78}{0.283} &\scalebox{0.78}{0.423} &\scalebox{0.78}{0.295} &\scalebox{0.78}{0.405} &\secondres{\scalebox{0.78}{0.279}} &\secondres{\scalebox{0.78}{0.404}} &\scalebox{0.78}{0.282} &\scalebox{0.78}{0.410} &\scalebox{0.78}{0.282} \\
\midrule
\multicolumn{2}{c|}{Weather} 
&\boldres{\scalebox{0.78}{0.228}} &\secondres{\scalebox{0.78}{0.268}} &\scalebox{0.78}{0.237} &\scalebox{0.78}{0.283} &\secondres{\scalebox{0.78}{0.229}} &\boldres{\scalebox{0.78}{0.265}} &\scalebox{0.78}{0.233} &\scalebox{0.78}{0.274} &\scalebox{0.78}{0.235} &\scalebox{0.78}{0.284} &\scalebox{0.78}{0.229} &\scalebox{0.78}{0.274} \\
    \bottomrule
  \end{tabular}
    \end{small}
  \end{threeparttable}
\end{table}

\begin{figure*}[t]
\centering
\includegraphics[width=6.5in]{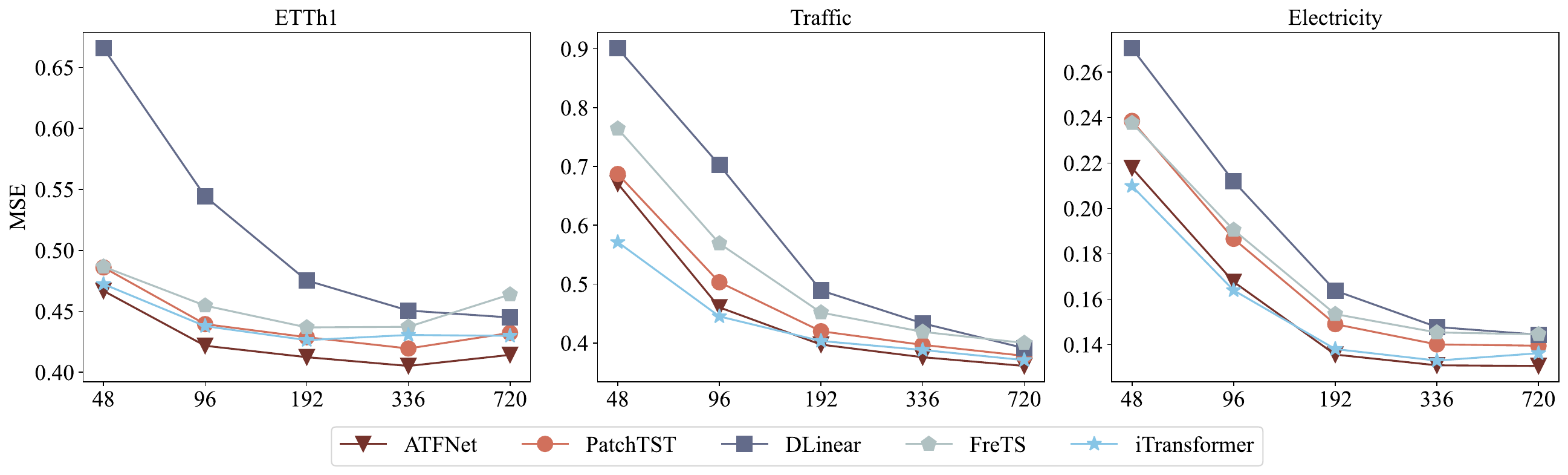}
\centering
\caption{Multi-variate long-term time series forecasting results (MSE) with 5 different look-back window size $L\in\{48, 96, 192, 336, 720\}$. We evaluate the performance of \model\ with 4 baseline models on ETTh1, Traffic and Electricity datasets. The prediction length is $T=96$. We also test on other prediction length $T\in\{192, 336, 720\}$. See Figure \ref{full_lookback} in Appendix for full results.}
\label{fig_lookback}
\end{figure*}

\textbf{Extended DFT.} From the results we can observe that when employing Extended DFT, \model\ can perform better than \model\ v1 with conventional DFT. This indicates that with the help of Extended DFT, the model can better capture the spectral similarity between  input and complete sequence.

\textbf{Dominant Harmonic Series Energy Weighting.} We can observe that \model\ achieves better performance than both T-Block and F-Block, which indicates that our proposed \model\ succeeds in combining 2 distinct models efficiently. Besides, \model\ outperforms \model\ v2, which utilizes a simple average stategy instead of Dominant Harmonic Series Energy Weighting. This proves that our proposed weighting method can allocate appropriate weights for time series with distinct periodic property and make full use of the advantages of both T-Block and F-Block.

\textbf{Complex-valued Network.} Comparing the performance of \model\ and \model\ v3, We find that using a real-valued network to approximate a complex-valued network leads to performance losses. Previous work~\citep{cao2020spectral} overlooked this aspect, resulting in sub-optimal results.

\subsection{Varying Look-back Window Size.}

Previous works~\cite{zeng2023transformers} have  noticed that longer input length cannot always contribute to better forecasting performance, especially for most of Transformer-based methods, since time points in too early stage may be less related to current and future temporal patterns, which can distract the attention and lead the models to more likely to overfit the training data.
Although our proposed F-Block is also based on attention mechanism, it can avoid such problem since the dependencies in frequency domain get less influence by longer historical input than time domain.

From the results shown in Figure \ref{fig_lookback}, it can be observed that the performance of our proposed \model\  consistently improves as the look-back window size increases. This finding suggests that our methods are capable of capturing more information from longer input sequences, which helps in avoiding overfitting and leads to enhanced performance. 

\section{Conclusion}

In this study, we introduce \model, a novel approach that innovatively integrates a frequency domain module with a time domain module for time series forecasting. This dual-module ensemble is designed to effectively capture both global and local dependencies inherent in time series data. A standout feature of \model\ is the Dominant Harmonic Series Energy Weighting mechanism, which intelligently adapts the contribution of each module based on the specific periodicity patterns of the time series. It also provides with a valuable way to measure the extent to which a time series exhibit periodicity, which is verified to be effective both theoretically and empirically. For the frequency module, we propose Extended DFT and Complex-valued Spectrum Attention, which play a crucial role in extracting essential frequency domain insights and enhancing the application of Transformer in the frequency domain. Extensive experiments conducted on various real-world datasets demonstrate that \model\ outperforms existing state-of-the-art time series forecasting methods. This advancement not only illustrates the efficacy of our model but also lays the groundwork for future explorations into the combined use of time and frequency domain analyses for enhanced forecasting accuracy. 

\clearpage

\bibliography{main}

\begin{thebibliography}{40}
\providecommand{\natexlab}[1]{#1}
\providecommand{\url}[1]{\texttt{#1}}
\expandafter\ifx\csname urlstyle\endcsname\relax
  \providecommand{\doi}[1]{doi: #1}\else
  \providecommand{\doi}{doi: \begingroup \urlstyle{rm}\Url}\fi

\bibitem[Bai et~al.(2018)Bai, Kolter, and Koltun]{bai2018empirical}
Bai, S., Kolter, J.~Z., and Koltun, V.
\newblock An empirical evaluation of generic convolutional and recurrent networks for sequence modeling.
\newblock \emph{arXiv preprint arXiv:1803.01271}, 2018.

\bibitem[Bassey et~al.(2021)Bassey, Qian, and Li]{bassey2021survey}
Bassey, J., Qian, L., and Li, X.
\newblock A survey of complex-valued neural networks.
\newblock \emph{arXiv preprint arXiv:2101.12249}, 2021.

\bibitem[Box \& Jenkins(1968)Box and Jenkins]{box1968some}
Box, G.~E. and Jenkins, G.~M.
\newblock Some recent advances in forecasting and control.
\newblock \emph{Journal of the Royal Statistical Society. Series C (Applied Statistics)}, 17\penalty0 (2):\penalty0 91--109, 1968.

\bibitem[Camacho \& Harris(2008)Camacho and Harris]{camacho2008sawtooth}
Camacho, A. and Harris, J.~G.
\newblock A sawtooth waveform inspired pitch estimator for speech and music.
\newblock \emph{The Journal of the Acoustical Society of America}, 124\penalty0 (3):\penalty0 1638--1652, 2008.

\bibitem[Cao et~al.(2020)Cao, Wang, Duan, Zhang, Zhu, Huang, Tong, Xu, Bai, Tong, et~al.]{cao2020spectral}
Cao, D., Wang, Y., Duan, J., Zhang, C., Zhu, X., Huang, C., Tong, Y., Xu, B., Bai, J., Tong, J., et~al.
\newblock Spectral temporal graph neural network for multivariate time-series forecasting.
\newblock \emph{Advances in neural information processing systems}, 33:\penalty0 17766--17778, 2020.

\bibitem[Challu et~al.(2022)Challu, Olivares, Oreshkin, Garza, Mergenthaler-Canseco, and Dubrawski]{challu2022nhits}
Challu, C., Olivares, K.~G., Oreshkin, B.~N., Garza, F., Mergenthaler-Canseco, M., and Dubrawski, A.
\newblock N-hits: Neural hierarchical interpolation for time series forecasting, 2022.

\bibitem[Chiheb et~al.(2017)Chiheb, Bilaniuk, Serdyuk, et~al.]{chiheb2017deep}
Chiheb, T., Bilaniuk, O., Serdyuk, D., et~al.
\newblock Deep complex networks.
\newblock In \emph{International Conference on Learning Representations}, 2017.

\bibitem[Chung et~al.(2014)Chung, Gulcehre, Cho, and Bengio]{chung2014empirical}
Chung, J., Gulcehre, C., Cho, K., and Bengio, Y.
\newblock Empirical evaluation of gated recurrent neural networks on sequence modeling.
\newblock \emph{arXiv preprint arXiv:1412.3555}, 2014.

\bibitem[De~Cheveign{\'e} \& Kawahara(2002)De~Cheveign{\'e} and Kawahara]{de2002yin}
De~Cheveign{\'e}, A. and Kawahara, H.
\newblock Yin, a fundamental frequency estimator for speech and music.
\newblock \emph{The Journal of the Acoustical Society of America}, 111\penalty0 (4):\penalty0 1917--1930, 2002.

\bibitem[Gfeller et~al.(2020)Gfeller, Frank, Roblek, Sharifi, Tagliasacchi, and Velimirovi{\'c}]{gfeller2020spice}
Gfeller, B., Frank, C., Roblek, D., Sharifi, M., Tagliasacchi, M., and Velimirovi{\'c}, M.
\newblock Spice: Self-supervised pitch estimation.
\newblock \emph{IEEE/ACM Transactions on Audio, Speech, and Language Processing}, 28:\penalty0 1118--1128, 2020.

\bibitem[Gough(1994)]{gough1994fast}
Gough, P.~T.
\newblock A fast spectral estimation algorithm based on the fft.
\newblock \emph{IEEE transactions on signal processing}, 42\penalty0 (6):\penalty0 1317--1322, 1994.

\bibitem[Hirose \& Yoshida(2012)Hirose and Yoshida]{hirose2012generalization}
Hirose, A. and Yoshida, S.
\newblock Generalization characteristics of complex-valued feedforward neural networks in relation to signal coherence.
\newblock \emph{IEEE Transactions on Neural Networks and learning systems}, 23\penalty0 (4):\penalty0 541--551, 2012.

\bibitem[Hochreiter \& Schmidhuber(1997)Hochreiter and Schmidhuber]{hochreiter1997long}
Hochreiter, S. and Schmidhuber, J.
\newblock Long short-term memory.
\newblock \emph{Neural computation}, 9\penalty0 (8):\penalty0 1735--1780, 1997.

\bibitem[Holt(2004)]{holt2004forecasting}
Holt, C.~C.
\newblock Forecasting seasonals and trends by exponentially weighted moving averages.
\newblock \emph{International journal of forecasting}, 20\penalty0 (1):\penalty0 5--10, 2004.

\bibitem[Kim et~al.(2018)Kim, Salamon, Li, and Bello]{kim2018crepe}
Kim, J.~W., Salamon, J., Li, P., and Bello, J.~P.
\newblock Crepe: A convolutional representation for pitch estimation.
\newblock In \emph{2018 IEEE International Conference on Acoustics, Speech and Signal Processing (ICASSP)}, pp.\  161--165. IEEE, 2018.

\bibitem[Kim et~al.(2021)Kim, Kim, Tae, Park, Choi, and Choo]{kim2021reversible}
Kim, T., Kim, J., Tae, Y., Park, C., Choi, J.-H., and Choo, J.
\newblock Reversible instance normalization for accurate time-series forecasting against distribution shift.
\newblock In \emph{International Conference on Learning Representations}, 2021.

\bibitem[Kingma \& Ba(2014)Kingma and Ba]{kingma2014adam}
Kingma, D.~P. and Ba, J.
\newblock Adam: A method for stochastic optimization.
\newblock \emph{arXiv preprint arXiv:1412.6980}, 2014.

\bibitem[Kitaev et~al.(2020)Kitaev, Kaiser, and Levskaya]{kitaev2020reformer}
Kitaev, N., Kaiser, {\L}., and Levskaya, A.
\newblock Reformer: The efficient transformer.
\newblock \emph{arXiv preprint arXiv:2001.04451}, 2020.

\bibitem[Lai et~al.(2018)Lai, Chang, Yang, and Liu]{lai2018modeling}
Lai, G., Chang, W.-C., Yang, Y., and Liu, H.
\newblock Modeling long-and short-term temporal patterns with deep neural networks.
\newblock In \emph{The 41st international ACM SIGIR conference on research \& development in information retrieval}, pp.\  95--104, 2018.

\bibitem[Liu et~al.(2022)Liu, Zeng, Chen, Xu, Lai, Ma, and Xu]{liu2022scinet}
Liu, M., Zeng, A., Chen, M., Xu, Z., Lai, Q., Ma, L., and Xu, Q.
\newblock Scinet: Time series modeling and forecasting with sample convolution and interaction.
\newblock \emph{Advances in Neural Information Processing Systems}, 35:\penalty0 5816--5828, 2022.

\bibitem[Liu et~al.(2023)Liu, Hu, Zhang, Wu, Wang, Ma, and Long]{liu2023itransformer}
Liu, Y., Hu, T., Zhang, H., Wu, H., Wang, S., Ma, L., and Long, M.
\newblock itransformer: Inverted transformers are effective for time series forecasting.
\newblock \emph{arXiv preprint arXiv:2310.06625}, 2023.

\bibitem[Luo et~al.(2016)Luo, Xie, and Xie]{luo2016interpolated}
Luo, J., Xie, Z., and Xie, M.
\newblock Interpolated dft algorithms with zero padding for classic windows.
\newblock \emph{Mechanical Systems and Signal Processing}, 70:\penalty0 1011--1025, 2016.

\bibitem[Nie et~al.(2022)Nie, Nguyen, Sinthong, and Kalagnanam]{nie2022time}
Nie, Y., Nguyen, N.~H., Sinthong, P., and Kalagnanam, J.
\newblock A time series is worth 64 words: Long-term forecasting with transformers.
\newblock In \emph{The Eleventh International Conference on Learning Representations}, 2022.

\bibitem[Oppenheim \& Verghese(2015)Oppenheim and Verghese]{Oppenheim2015Signal}
Oppenheim and Verghese.
\newblock \emph{Signals, Systems, and Inference}.
\newblock Pearson, 2015.
\newblock ISBN 9780133944211.

\bibitem[Oreshkin et~al.(2019)Oreshkin, Carpov, Chapados, and Bengio]{oreshkin2019n}
Oreshkin, B.~N., Carpov, D., Chapados, N., and Bengio, Y.
\newblock N-beats: Neural basis expansion analysis for interpretable time series forecasting.
\newblock \emph{arXiv preprint arXiv:1905.10437}, 2019.

\bibitem[Parseval \& Mémoire(1806)Parseval and Mémoire]{Parseval}
Parseval and Mémoire, M.-A.
\newblock Sur les séries et sur l'intégration complète d'une équation aux différences partielles linéaire du second ordre, à coefficients constants.
\newblock \emph{Sciences, mathématiques et physiques}, pp.\  638--648, 1806.

\bibitem[Paszke et~al.(2019)Paszke, Gross, Massa, Lerer, Bradbury, Chanan, Killeen, Lin, Gimelshein, Antiga, et~al.]{paszke2019pytorch}
Paszke, A., Gross, S., Massa, F., Lerer, A., Bradbury, J., Chanan, G., Killeen, T., Lin, Z., Gimelshein, N., Antiga, L., et~al.
\newblock Pytorch: An imperative style, high-performance deep learning library.
\newblock \emph{Advances in neural information processing systems}, 32, 2019.

\bibitem[Santamaria et~al.(2000)Santamaria, Pantaleon, and Ibanez]{santamaria2000comparative}
Santamaria, I., Pantaleon, C., and Ibanez, J.
\newblock A comparative study of high-accuracy frequency estimation methods.
\newblock \emph{Mechanical Systems and Signal Processing}, 14\penalty0 (5):\penalty0 819--834, 2000.

\bibitem[Sun \& Boning(2022)Sun and Boning]{sun2022fredo}
Sun, F.-K. and Boning, D.~S.
\newblock Fredo: Frequency domain-based long-term time series forecasting.
\newblock \emph{arXiv e-prints}, pp.\  arXiv--2205, 2022.

\bibitem[Vaswani et~al.(2017)Vaswani, Shazeer, Parmar, Uszkoreit, Jones, Gomez, Kaiser, and Polosukhin]{vaswani2017attention}
Vaswani, A., Shazeer, N., Parmar, N., Uszkoreit, J., Jones, L., Gomez, A.~N., Kaiser, {\L}., and Polosukhin, I.
\newblock Attention is all you need.
\newblock \emph{Advances in neural information processing systems}, 30, 2017.

\bibitem[Wu et~al.(2021)Wu, Xu, Wang, and Long]{wu2021autoformer}
Wu, H., Xu, J., Wang, J., and Long, M.
\newblock Autoformer: Decomposition transformers with auto-correlation for long-term series forecasting.
\newblock \emph{Advances in Neural Information Processing Systems}, 34:\penalty0 22419--22430, 2021.

\bibitem[Wu et~al.(2022)Wu, Hu, Liu, Zhou, Wang, and Long]{wu2022timesnet}
Wu, H., Hu, T., Liu, Y., Zhou, H., Wang, J., and Long, M.
\newblock Timesnet: Temporal 2d-variation modeling for general time series analysis.
\newblock In \emph{The Eleventh International Conference on Learning Representations}, 2022.

\bibitem[Xu et~al.(2023)Xu, Zeng, and Xu]{xu2023fits}
Xu, Z., Zeng, A., and Xu, Q.
\newblock Fits: Modeling time series with $10 k $ parameters.
\newblock \emph{arXiv preprint arXiv:2307.03756}, 2023.

\bibitem[Xu et~al.(2024)Xu, Zeng, and Xu]{xu2024fits}
Xu, Z., Zeng, A., and Xu, Q.
\newblock Fits: Modeling time series with $10k$ parameters, 2024.

\bibitem[Yi et~al.(2023)Yi, Zhang, Fan, Wang, Wang, He, An, Lian, Cao, and Niu]{yi2023frequency}
Yi, K., Zhang, Q., Fan, W., Wang, S., Wang, P., He, H., An, N., Lian, D., Cao, L., and Niu, Z.
\newblock Frequency-domain mlps are more effective learners in time series forecasting.
\newblock In \emph{Thirty-seventh Conference on Neural Information Processing Systems}, 2023.

\bibitem[Zeng et~al.(2023)Zeng, Chen, Zhang, and Xu]{zeng2023transformers}
Zeng, A., Chen, M., Zhang, L., and Xu, Q.
\newblock Are transformers effective for time series forecasting?
\newblock In \emph{Proceedings of the AAAI conference on artificial intelligence}, volume~37, pp.\  11121--11128, 2023.

\bibitem[Zhang \& Yan(2022)Zhang and Yan]{zhang2022crossformer}
Zhang, Y. and Yan, J.
\newblock Crossformer: Transformer utilizing cross-dimension dependency for multivariate time series forecasting.
\newblock In \emph{The Eleventh International Conference on Learning Representations}, 2022.

\bibitem[Zhou et~al.(2021)Zhou, Zhang, Peng, Zhang, Li, Xiong, and Zhang]{zhou2021informer}
Zhou, H., Zhang, S., Peng, J., Zhang, S., Li, J., Xiong, H., and Zhang, W.
\newblock Informer: Beyond efficient transformer for long sequence time-series forecasting.
\newblock In \emph{Proceedings of the AAAI conference on artificial intelligence}, volume~35, pp.\  11106--11115, 2021.

\bibitem[Zhou et~al.(2022{\natexlab{a}})Zhou, Ma, Wen, Sun, Yao, Yin, Jin, et~al.]{zhou2022film}
Zhou, T., Ma, Z., Wen, Q., Sun, L., Yao, T., Yin, W., Jin, R., et~al.
\newblock Film: Frequency improved legendre memory model for long-term time series forecasting.
\newblock \emph{Advances in Neural Information Processing Systems}, 35:\penalty0 12677--12690, 2022{\natexlab{a}}.

\bibitem[Zhou et~al.(2022{\natexlab{b}})Zhou, Ma, Wen, Wang, Sun, and Jin]{zhou2022fedformer}
Zhou, T., Ma, Z., Wen, Q., Wang, X., Sun, L., and Jin, R.
\newblock Fedformer: Frequency enhanced decomposed transformer for long-term series forecasting.
\newblock In \emph{International Conference on Machine Learning}, pp.\  27268--27286. PMLR, 2022{\natexlab{b}}.

\end{thebibliography}
\bibliographystyle{icml2024}

\newpage
\onecolumn
\appendix
\section{Proofs}
\subsection{Theorem 1}
\label{proof_theorem1}
\textbf{Theorem 1.} \textit{Let $F \in \mathbb{C}^{L+T}$ represent the output spectrum of the input series $X \in \mathbb{R}^L$ obtained through the Extended DFT, where $L$ denotes the input length and $T$ denotes the prediction length. We have the following statement holds
\begin{equation}
\begin{aligned}
\operatorname{Re}(F[k])&=\operatorname{Re}(F[L+T-k]),\\
\operatorname{Im}(F[k])&=-\operatorname{Im}(F[L+T-k]),  
\end{aligned}
\end{equation}
which implies that $F$ exhibits conjugate symmetry. Here, $k=[1,2,\dots,L+T-1]$, $\operatorname{Re}(\cdot)$ represents the real part, and $\operatorname{Im}(\cdot)$ represents the imaginary part.}

\begin{proof} 
Following Eq. \ref{eDFT}, we have
\begin{equation}
\begin{aligned}
    F[k]&=\sum_{n=0}^{L-1}x[n]e^{-2\pi i\frac{kn}{L+T}},\\
\end{aligned}
\end{equation}
\begin{equation}
    \begin{aligned}
        F[L+T-k]&=\sum_{n=0}^{L-1}x[n]e^{-2\pi i\frac{(L+T-k)n}{L+T}}\\
        &=\sum_{n=0}^{L-1}x[n]e^{-2\pi in}\cdot e^{2\pi i\frac{kn}{L+T}}\\
        &=\sum_{n=0}^{L-1}x[n]e^{2\pi i\frac{kn}{L+T}}.\\
    \end{aligned}
\end{equation}
Applying Euler's Formula, we obtain
\begin{equation}
    \operatorname{Re}(F[k])=\operatorname{Re}(F[L+T-k])=\sum_{n=0}^{L-1}x[n]\operatorname{cos}(2\pi \frac{kn}{L+T}),
\end{equation}
\begin{equation}
    \operatorname{Im}(F[k])=-\operatorname{Im}(F[L+T-k])=-\sum_{n=0}^{L-1}x[n]\operatorname{sin}(2\pi \frac{kn}{L+T}).
\end{equation}
Thus, we complete the proof of Theorem \ref{theorem:1}.
\end{proof}
\subsection{Theorem 2}
\label{proof_theorem2}
\textbf{Theorem 2.} \textit{Let $f(x)$ denote a continuous time series defined on the interval $[0,L]$. We denote the discrete form of length-$L$ realizations of $f(x)$, sampled at an interval of 1, as $X = [f(0), f(1), \dots, f(L-1)]$. $f(x)$ can be decomposed into two parts: $f(x)=f_p(x)+f_r(x)$, where $f_p(x)$ is a strictly periodic function with a period of $\tau$ (assuming $L=k\tau$ and $L, k, \tau\in \mathbb{N^*}$), i.e. $f_p(x-\tau)=f_p(x), \forall x\in[\tau, L]$. Let $E_p=\sum_{x=0}^{L-1}f_p^2(x) $ denotes the energy of periodic part and $E_r=\sum_{x=0}^{L-1}f_r^2(x)$ denotes the energy of the remaining part. 
Let $\lambda = E_p/E_r$, where $\lambda$ represents the magnitude of the periodic signal compared to the remaining signal, and varies with different dividing methods. 
Considering the frequency resolution limitations imposed by the sampling interval, we assume that the energy in the frequency component above the Nyquist frequency ($\pi$) is small enough to be neglected.
For simplicity, we additionally assume $\sum_{x=0}^{L-1} f(x)=0$. Under these assumptions, we have}
\begin{equation}
    \frac{E_h}{E_f}=\frac{\sum_{n=0}^{T-1} |F[nk]|^2}{\sum_{n=0}^{L-1} |F[n]|^2 }\geq \frac{\lambda-2\sqrt{\lambda}}{\lambda - 2\sqrt{\lambda} + 1},
\end{equation}
\textit{where $E_h$ represents the total energy of the dominant harmonic group with a basis frequency $k=L/\tau$, $E_f$ represents the total energy of the whole spectrum, and $F(\cdot)$ represents the DFT components at a specific frequency.}
\begin{proof}
    Since $f_p(x)$ is strictly periodic, it can be decomposed into its Fourier series. Based on the assumption that the energy in the frequency component above the Nyquist frequency ($\pi$) is small enough to be neglected, we decompose $f_p(x)$ into the first $q=\lceil \tau/2\rceil-1$ Fourier series as follows:
    \begin{equation} \label{eq:pf_fourier}
        f_p(x)=\sum_{m=-q}^{q}c_me^{2\pi i\frac{m}{\tau}x}.
    \end{equation}
    where the Fourier coefficients $c_m$ are determined through the integral:
    \begin{equation}
        c_m=\frac{1}{\tau}\int_0^{\tau}f(x)e^{2\pi i\frac{m}{\tau}x}dx.
    \end{equation}
    
    The energy of $f_p(x)$ is then computed using
    \begin{equation} 
    \begin{aligned} \label{eq:pf_energy_p_1}
        E_p &= \sum_{x=0}^{L-1}\Big[\sum_{m=-q}^{q}c_me^{2\pi i\frac{m}{\tau}x}\Big]^2 \\
        &= k\sum_{x=0}^{\tau-1}\Big[\sum_{m=-q}^{q}c_me^{2\pi i\frac{m}{\tau}x}\Big]^2 \\
        &= k\sum_{x=0}^{\tau-1}\sum_{m=-q}^{q}\sum_{n=-q}^{q}c_mc_ne^{2\pi i\frac{m+n}{\tau}x}\\
        &= k\sum_{m=-q}^{q}\sum_{n=-q}^{q}c_mc_n\sum_{x=0}^{\tau-1}e^{2\pi i\frac{m+n}{\tau}x} 
    \end{aligned}
    \end{equation}
    When $m+n\neq 0$, we have
    \begin{equation}
    \label{neq}
        \sum_{x=0}^{\tau-1}e^{2\pi i\frac{m+n}{\tau}x}=\frac{1-e^{2\pi i(m+n)}}{1-e^{2\pi i\frac{m+n}{\tau}}}=0.
    \end{equation}
    When $m+n= 0$, we have
    \begin{equation}
    \label{equal}
        \sum_{x=0}^{\tau-1}e^{2\pi i\frac{m+n}{\tau}x}=\tau.
    \end{equation}
    Combining Eq.\ref{eq:pf_energy_p_1}, Eq.\ref{neq} and Eq.\ref{equal}, we have
    \begin{equation}
        \begin{aligned}\label{eq:pf_energy_p}
            E_p &= k\sum_{x=0}^{\tau-1}\sum_{m=-q}^{q}\sum_{n=-q}^{q}c_mc_ne^{2\pi i\frac{m+n}{\tau}x}\\
            &= k \sum_{m=-q}^{q} \tau|c_m|^2\\
            &= L \sum_{m=-q}^{q} |c_m|^2
        \end{aligned}
    \end{equation}
    
    Similarly, the energy of $f(x)$ is given by
    \begin{equation}
        \begin{aligned}
            E_t &= \sum_{x=0}^{L-1}\big[\sum_{m=-q}^{q}c_me^{2\pi i\frac{m}{\tau}x} + f_r(x)\big]^2 \\
            &= \sum_{x=0}^{L-1}\big[\sum_{m=-q}^{q}c_me^{2\pi i\frac{m}{\tau}x}\big]^2 + \sum_{x=0}^{L-1}f_r^2(x) \\ 
            &+ 2\sum_{x=0}^{L-1}\sum_{m=-q}^{q}c_me^{2\pi i\frac{m}{\tau}x}f_r(x) \\
            &= E_p + E_r + 2\sum_{x=0}^{L-1}\sum_{m=-q}^{q}c_me^{2\pi i\frac{m}{\tau}x}f_r(x).
        \end{aligned}
    \end{equation}
    For the Fourier component of basis frequency $k=L/\tau$, we have
    \begin{equation} \label{eq:pf_basisfrequency1}
    \begin{aligned}
        F[k] &= \sum_{x=0}^{L-1}[\sum_{m=-q}^{q}c_me^{2\pi i\frac{m}{\tau}x} + f_r(x)]e^{-2\pi i\frac{k}{L}x}\\
        &= c_1L + \sum_{x=0}^{L-1}f_r(x)e^{-2\pi i\frac{1}{\tau}x},
    \end{aligned}
    \end{equation}
    where the last equality follows from performing simple geometric sequence summation. Similarly, the Fourier component of the remaining harmonics $nk (n=2, \dots, \tau -1)$ is computed as:
    \begin{equation} \label{eq:pf_basisfrequency2}
        F[nk] = c_nL + \sum_{x=0}^{L-1}f_r(x)e^{-2\pi i\frac{n}{\tau}x}.
    \end{equation}
    Combining Eq. \ref{eq:pf_energy_p}, Eq. \ref{eq:pf_basisfrequency1}, and Eq. \ref{eq:pf_basisfrequency2}, we achieve a lower bound of the total energy of this dominant harmonic series as follows
    \begin{equation} \label{eq:energyh}
        \begin{aligned}
            E_h = \sum_{n=0}^{\tau-1}|F[n k]|^2 &= \sum_{n=-q}^{q}|F[n k]|^2 \\
            &= \sum_{n=-q}^{q} |c_nL + \sum_{x=0}^{L-1}f_r(x)e^{-2\pi i\frac{n}{\tau}x}|^2\\
            &= L^2\sum_{n=-q}^{q}|c_n|^2 + 2L\sum_{x=0}^{L-1}\sum_{n=-q}^{q}c_ne^{2\pi i\frac{n}{\tau}x}f_r(x)
            + \sum_{n=-q}^{q}\big|\sum_{x=0}^{L-1}f_r(x)e^{-2\pi i\frac{n}{\tau}x} \big|^2\\
            &\geq L[E_p + 2\sum_{x=0}^{L-1}\sum_{n=-q}^{q}c_ne^{2\pi i\frac{n}{\tau}x}f_r(x)],
        \end{aligned}
    \end{equation}
    
    Applying Parseval's theorem \cite{Parseval}, we have
    \begin{equation} \label{eq:parseval}
        E_f = LE_t = L[E_p + E_r + 2\sum_{x=0}^{L-1}\sum_{m=-q}^{q}c_me^{2\pi i\frac{m}{\tau}x}f_r(x)].
    \end{equation}
    Consequently, by Eq. \ref{eq:energyh} and Eq. \ref{eq:parseval}, we can decuce that
    \begin{equation} \label{eq:ratio}
        \frac{E_h}{E_f} \geq \frac{E_p + 2\sum_{x=0}^{L-1}\sum_{m=-q}^{q}c_me^{2\pi i\frac{m}{\tau}x}f_r(x)}{E_p + E_r + 2\sum_{x=0}^{L-1}\sum_{m=-q}^{q}c_me^{2\pi i\frac{m}{\tau}x}f_r(x)}.
    \end{equation}
    Applying the Cauchy–Schwarz inequality, we obtain
    \begin{equation}
    \begin{aligned}
        \big[\sum_{x=0}^{L-1}\sum_{m=-q}^{q}c_me^{2\pi i\frac{m}{\tau}x}f_r(x)\big]^2 &\leq\sum_{x=0}^{L-1}[\sum_{m=-q}^{q}c_me^{2\pi i\frac{m}{\tau}x}]^2\sum_{x=0}^{L-1}f_r^2(x) \\
        & = E_p E_r,
    \end{aligned}
    \end{equation}
    where the last equality follows from Eq. \ref{eq:pf_fourier} and the definition of $E_p$ and $E_r$. Therefore, we have
    \begin{equation} \label{eq:bound}
        -\sqrt{E_pE_r} \leq \sum_{x=0}^{L-1}\sum_{m=-q}^{q}c_me^{2\pi i\frac{m}{\tau}x}f_r(x) \leq \sqrt{E_pE_r}.
    \end{equation}
    Combining Eq. \ref{eq:ratio} and Eq. \ref{eq:bound}, we obtain
    \begin{equation}
        \frac{E_h}{E_f} \geq\frac{E_p-2\sqrt{E_pE_r}}{E_p+E_r-2\sqrt{E_pE_r}} = \frac{\lambda-2\sqrt{\lambda}}{\lambda - 2\sqrt{\lambda} + 1}.
    \end{equation}
    Thus, we complete the proof of Theorem \ref{theorem:2}.
\end{proof}

\section{Discussion on Complex-Valued Neural Network}
\label{cvnn}
Our model employs complex-valued neural network (CVNN) structures, with parameters defined on the complex plane. Several studies in frequency domain time series forecasting method mentioned in Section \ref{sec:related_work} have also naturally employed such structures, owing to the complex number format of spectrums derived from FFT. However, these studies provide little analysis for the effectiveness of CVNNs. It's important to note that while transforming complex number into real numbers for application in real-valued neural networks (RVNNs) is feasible, CVNNs offer superior capability in capturing complex representations within the frequency domain. 

Our analysis begins with a complex linear layer that maps $w \in \mathbb{C}^{d_1}$ to $z=\boldsymbol{B}w\in \mathbb{C}^{d_2}$, where $\boldsymbol{B} \in \mathbb{C}^{d_2 \times d_1}$. The bias term can be integrated into $\boldsymbol{B}$ by appending a scaler $1$ to $w$. A common RVNN approach, such as that used in \citet{cao2020spectral}, treats the real and imaginary parts of $w$ as distinct entities, $w_r \in \mathbb{R}^{d_1}$ and $w_c \in \mathbb{R}^{d_2}$. It employs two real weight matrices $\boldsymbol{B}_r, \boldsymbol{B}_c \in \mathbb{R}^{d_2 \times d_1}$, resulting in $z= \boldsymbol{B}_r w_r + i\boldsymbol{B}_c w_c$. This method relies solely on the real (imaginary) component of the input to produce the real (imaginary) component of the output. We use a scenario where $d_1=2$ with one dimension representing the bias and $d_2=1$ to show that the method is inadequate for representing even such simple transformations. 

Consider a complex linear function $f(w)=\beta_0+\beta_1 z$ with input $w\in\mathbb{C}$ and coefficients $\beta_0,\beta_1\in\mathbb{C}$. Now we have observations $\{(w_j,z_j)\}_{j=1}^{n}$ with random noise, where $z_j = f(w_j) + \epsilon_j$. 
We can write the observations into the following matrix form:
\begin{equation}
    \left[\begin{array}{c}
z_1 \\
z_2 \\
\vdots\\
z_n
\end{array}\right] = \bz = \bW\left[\begin{array}{c}
\beta_0 \\
\beta_1
\end{array}\right] + \bE =  \left[\begin{array}{cc}
1 & w_1 \\
1 & w_2 \\
\vdots& \vdots\\
1 & w_n
\end{array}\right] \left[\begin{array}{c}
\beta_0 \\
\beta_1
\end{array}\right] + \left[\begin{array}{c}
\epsilon_1 \\
\epsilon_2 \\
\vdots \\
\epsilon_n
\end{array}\right],
\end{equation}
where $\bz,\bE\in\mathbb{C}^{n\times1}, \bW\in\mathbb{C}^{n\times2}$. First, we can estimate the model directly using the complex method:
\begin{equation}
    \hat{\beta}_0,\hat{\beta}_1 = \underset{\beta_0,\beta_1\in\mathbb{C}}{\operatorname{argmin}}\left\|\bz-\bW \left[\begin{array}{c}
\beta_0 \\
\beta_1
\end{array}\right] \right\|_2^2.
\end{equation}
By a few steps of differentiation, we derive
\begin{equation}
\left[\begin{array}{c}
\hat{\beta}_0 \\
\hat{\beta}_1
\end{array}\right] = \left(\bW^* \bW\right)^{-1} \bW^*\bz,
\end{equation}
where $\bW^{*}=\Bar{\bW}^T$ represents conjugate transpose of $\bW$. Then we have
\begin{equation}
\begin{aligned}
\underset{{n \rightarrow \infty}}{\operatorname{plim}}\left[\begin{array}{c}
\hat{\beta}_0 \\
\hat{\beta}_1
\end{array}\right] &= \underset{{n \rightarrow \infty}}{\operatorname{plim}}\left[ \left(\bW^* \bW\right)^{-1} \bW^*(\bW \left[\begin{array}{c}
\beta_0 \\
\beta_1
\end{array}\right] + \bE) \right] \\
&=\left[\begin{array}{c}
\beta_0 \\
\beta_1
\end{array}\right] + \boldsymbol{Q}\underset{{n \rightarrow \infty}}{\operatorname{plim}}\left[ \bW^*\bE/n \right] =\left[\begin{array}{c}
\beta_0 \\
\beta_1
\end{array}\right],
\end{aligned}
\end{equation}
where $\boldsymbol{Q}=\operatorname{plim}_{{n \rightarrow \infty}}$ is the limit operator in probability. Here, $\boldsymbol{Q}=\operatorname{plim}_{{n \rightarrow \infty}}\bW^*\bW/n$ is assumed to be rank-2 matrix with finite elements, and we make another reasonable assumption $E(\bW^*\bE)=\mathbf{0}$ which ensures the last equality holds. Hence, the complex method is capable of obtaining coefficients that converge to the true parameters. 

Conversely, consider the real method that handles the real and imaginary parts of $\bW$ independently. We demonstrate that this method fails to ensure convergence to the true parameters. We write all the complex number in real format as follows
\begin{equation}
\begin{aligned}
    w_j &= u_j + iv_j\\
    \epsilon_j &= \epsilon_j^r + i\epsilon_j^i,\\
    \beta_0 &= \beta_0^r + i\beta_0^i,\quad \beta_1 = \beta_1^r + i\beta_1^i,
\end{aligned}
\end{equation}
where $u_j,v_j,\epsilon_j^r,\epsilon_j^i,\beta_0^r,\beta_0^i,\beta_1^i,\beta_1^i\in\mathbb{R}$. Applying these to the true model formulation, we have $x_j\in\mathbb{R}$ and $y_j\in\mathbb{R}$ take the forms
\begin{equation}
\begin{aligned} \label{eq:pf_real1}
    x_j &= \beta_0^r + \beta_1^ru_j - \beta_1^iv_j + \epsilon_j^r \\
    y_j &= \beta_0^i + \beta_1^iu_j + \beta_1^rv_j + \epsilon_j^i,
\end{aligned}  
\end{equation}
where $z_j= x_j + iy_j$. If we only use $u_j$ to estimate $x_j$ and only use $v_j$ to estiamte $y_j$, as shown below
\begin{equation}
\begin{aligned} \label{eq:pf_real2}
    \hat{x_j} &= \hat{\gamma}_0 + \hat{\gamma}_1u_j,\\
    \hat{y_j} &= \hat{\delta}_0 + \hat{\delta}_1v_j,
\end{aligned}  
\end{equation}
this obviously leads to a loss of information. The estimates are obtained by
\begin{equation}
\begin{aligned}
    \hat{\gamma}_0,\hat{\gamma}_1 &= \underset{\gamma_0,\gamma_1\in\mathbb{R}}{\operatorname{argmin}}\left\|\bx-\bU \left[\begin{array}{c}
\gamma_0 \\
\gamma_1
\end{array}\right] \right\|_2^2,\\
    \hat{\delta}_0,\hat{\delta}_1 &= \underset{\delta_0,\delta_1\in\mathbb{R}}{\operatorname{argmin}}\left\|\by-\bV \left[\begin{array}{c}
\delta_0 \\
\delta_1
\end{array}\right] \right\|_2^2,
\end{aligned}
\end{equation}
where 
\begin{equation}
\bx=\left[\begin{array}{c}
x_1 \\
x_2 \\
\vdots\\
x_n
\end{array}\right]\in\mathbb{R}^{n\times1},\quad \by=\left[\begin{array}{c}
y_1 \\
y_2 \\
\vdots\\
y_n
\end{array}\right]\in\mathbb{R}^{n\times1},\quad \bU = \left[\begin{array}{cc}
1 & u_1 \\
1 & u_2 \\
\vdots& \vdots\\
1 & u_n
\end{array}\right]\in\mathbb{R}^{n\times2},\quad \bV=\left[\begin{array}{cc}
1 & v_1 \\
1 & v_2 \\
\vdots& \vdots\\
1 & v_n
\end{array}\right]\in\mathbb{R}^{n\times2}.
\end{equation}
By simple derivations, we obtain
\begin{equation}
\begin{aligned} \label{eq:pf_real4}
\left[\begin{array}{c}
\hat{\gamma}_0 \\
\hat{\gamma}_1
\end{array}\right] &= \left[\begin{array}{c}
\beta_0^r \\
\beta_1^r
\end{array}\right]-(\bU^{T}\bU)\bU^{T}(\bV\left[\begin{array}{c}
0 \\
\beta_1^i
\end{array}\right]+\bE^r),\\
\left[\begin{array}{c}
\hat{\delta}_0 \\
\hat{\delta}_1
\end{array}\right] &= \left[\begin{array}{c}
\beta_0^i \\
\beta_1^r
\end{array}\right]+(\bV^{T}\bV)\bV^{T}(\bU\left[\begin{array}{c}
0 \\
\beta_1^i
\end{array}\right]+\bE^i),
\end{aligned}.
\end{equation}
Applying the probability limit operator to Eq. \ref{eq:pf_real4} and assuming $\boldsymbol{Q}^r=\operatorname{plim}_{{n \rightarrow \infty}}\bU^*\bU/n, \boldsymbol{Q}^i=\operatorname{plim}_{{n \rightarrow \infty}}\bV^*\bV/n$ are rank-2 matrices with finite elements and $E(\bU^T\bE^r)=E(\bV^T\bE^i)=\mathbf{0}$, we have

\begin{equation}
\begin{aligned} \label{eq:pf_real3}
\underset{{n \rightarrow \infty}}{\operatorname{plim}}\left[\begin{array}{c}
\hat{\gamma}_0 \\
\hat{\gamma}_1
\end{array}\right] &= \left[\begin{array}{c}
\beta_0^r \\
\beta_1^r
\end{array}\right] - \boldsymbol{Q}^r\underset{{n \rightarrow \infty}}{\operatorname{plim}}\bU^{T}\bV\left[\begin{array}{c}
0 \\
\beta_1^i
\end{array}\right]\Big/n \neq \left[\begin{array}{c}
\beta_0^r \\
\beta_1^r
\end{array}\right], \\
\underset{{n \rightarrow \infty}}{\operatorname{plim}}\left[\begin{array}{c}
\hat{\delta}_0 \\
\hat{\delta}_1
\end{array}\right] &= \left[\begin{array}{c}
\beta_0^i \\
\beta_1^r
\end{array}\right]+\boldsymbol{Q}^i\underset{{n \rightarrow \infty}}{\operatorname{plim}}\bV^{T}\bU\left[\begin{array}{c}
0 \\
\beta_1^i
\end{array}\right]\Big/n \neq \left[\begin{array}{c}
\beta_0^i \\
\beta_1^r
\end{array}\right].
\end{aligned}
\end{equation}
Therefore, from Eq. \ref{eq:pf_real1}, Eq. \ref{eq:pf_real2}, and Eq. \ref{eq:pf_real3}, we show that this method do not ensure convergence in probability. These derivations can be easily generalize to scenarios with arbitrary $d_1$ and $d_2$, thus reflecting the limitations of this RVNN method in representing complex transformations.

Another RVNN methodology concatenate the real and imaginary parts into a single real vector $\Tilde{w}=(w_r^{T},w_c^{T})^{T} \in \mathbb{R}^{2d_1}$ and uses a single real weight matrix $\Tilde{\boldsymbol{B}} \in \mathbb{R}^{2d_1 \times 2d_2}$ to produce an $2d_2$-dimensional vector $\Tilde{z}=\Tilde{\boldsymbol{B}}\Tilde{w}$. The first $d_2$ elements of $\Tilde{z}$ correspond to the real part of the output vector and the remaining correspond to imaginary part, i.e.,$z=\Tilde{z}[:d_2]+i\Tilde{z}[d_2:]$. While this formulation can technically represent any complex linear layer $W$ mentioned above, it increases the complexity of the network by doubling the number of parameters compared to the CVNN approach. This not only makes the learning process more challenging but also overlooks the intrinsic properties of complex number such as phase rotation and amplitude modulation. This aspect of RVNNs, along with the inherent advantages of CVNNs in handling complex number characteristics, is discussed in detail in \citet{hirose2012generalization,chiheb2017deep,bassey2021survey}.

\section{Discussion on Fundamental Frequency Detection}
\label{detection}
Detecting the fundamental frequency constitutes a crucial aspect of Dominant Harmonic Group Energy Weighting. As discussed in Section \ref{dominant}, we can adopt either a straightforward approach that directly selects the frequency with the highest amplitude or refer to pitch detection algorithms documented in the audio processing literature to address this issue. It should be noted that the accuracy of fundamental frequency detection algorithms may vary depending on the characteristics of the time series, thereby posing a challenge in selecting an optimal choice. Through our investigation, we have discovered that the naive method surpasses other pitch detection algorithms in terms of both accuracy and computational efficiency, leading us to designate it as the default fundamental frequency detection algorithm.

Furthermore, we conducted experiments to assess the performance of distinct fundamental frequency detection algorithms, including YIN \cite{de2002yin} and SWIPE \cite{camacho2008sawtooth}. Specifically, we employed a univariate time series exhibiting strong periodicity and divided it into several time frames of varying lengths. Our evaluation aimed to determine whether these algorithms accurately detected the fundamental frequency and to quantify their computational costs. We utilized the Traffic dataset, with the variable denoted as '8' and spanning indices 0-3000, exhibiting an evident period of 24. For each length, we sampled 2000 time frames using a sliding window with a stride of 1. The corresponding results are presented in Table \ref{table_fund}.


\begin{table}[htbp]
  \caption{Results of distinct fundamental frequency detection algorithms.}
  \label{table_fund}
  \vskip 0.05in
  \centering
  \begin{threeparttable}
  \begin{small}
  \renewcommand{\multirowsetup}{\centering}
  \setlength{\tabcolsep}{4.5pt}
  \begin{tabular}{c|cc|cc|cc|cc}
    \toprule
     Time Frame Length & 
    \multicolumn{2}{c}{96} &
    \multicolumn{2}{c}{192} &
    \multicolumn{2}{c}{336} &
    \multicolumn{2}{c}{720}  \\
    \cmidrule(lr){2-3}\cmidrule(lr){4-5}\cmidrule(lr){6-7}\cmidrule(lr){8-9}
    & Acc & Time(s) & Acc & Time(s) & Acc & Time(s) & Acc & Time(s) \\
    
    \midrule

Naive & 100.0\% & 0.021 & 100.0\% & 0.020 & 100.0\% & 0.024 & 100.0\% & 0.031  \\
\midrule
YIN & 88.8\% & 2.863 & 97.3\% & 10.909 & 99.2\% & 32.659 & 100.0\% & 148.357  \\
\midrule
SWIPE & 97.2\% & 0.446 & 98.0\% & 1.043 & 100.0\% & 1.974 & 100.0\% & 4.935 \\

    \bottomrule
  \end{tabular}
    \end{small}
  \end{threeparttable}
\end{table}

The results clearly demonstrate that the naive method, which selects the frequency with the highest amplitude as the fundamental frequency, outperforms other algorithms in terms of both accuracy and computational efficiency.

\section{Experiments Details}
\subsection{Datasets}
We include 8 real-world datasets that are widely used for evaluation of long-term time series forecasting.

\textbf{ETT:} The ETT(Electricity Transformer Temperature) datasets are collected from two different electric transformers labeled with 1 and 2, recording loads and oil temperature of transformers. We employ 4 ETT sub-datasets: ETTh1, ETTh2, ETTm1, ETTm2 in different resolutions (\textit{h} for 1-hour and \textit{m} for 15-min).

\textbf{Electricity:} It records the hourly electricity consumption of 321 clients from 2012 to 2014.

\textbf{Traffic:} It records the road occupancy rates measured by different sensors on San Francisco Bay area freeways, provided by California Department of Transportation.

\textbf{Weather:} It collects 21 meteorological indicators, such as humidity and air temperature, from the Weather Station of the Max Planck Biogeochemistry Institute in Germany in 2020.

\textbf{Solar:} The Solar Energy dataset records the solar power production of 137 PV plants in 2006.

The detailed dataset descriptions are included in Table \ref{dataset}.
\begin{table}[htbp]
  \caption{Dataset descriptions. \textit{Dim} stands for the number of variate in the dataset. \textit{Time Points} represents the total number of time points. \textit{Frequency} denotes the sampling interval of time points. \textit{Information} denotes the domain that the dataset is from.}
  \label{dataset}
  \vskip 0.05in
  \centering
  \begin{threeparttable}
  \begin{small}
  \renewcommand{\multirowsetup}{\centering}
  \setlength{\tabcolsep}{4.5pt}
  \begin{tabular}{c|c|c|c|cc}
    \toprule
    {Dataset} & 
    {\rotatebox{0}{\scalebox{0.8}{Dim}}} &
    {\rotatebox{0}{\scalebox{0.8}{{Time Points}}}} &
    {\rotatebox{0}{\scalebox{0.8}{Frequency}}} &
    {\rotatebox{0}{\scalebox{0.8}{Information}}} & \\
    \midrule
{{\scalebox{0.95}{ETTh1}}} 
&\scalebox{0.78}{7} &\scalebox{0.78}{17420} &\scalebox{0.78}{1-hour} &\scalebox{0.78}{Electricity}  \\
\midrule
{{\scalebox{0.95}{ETTh2}}} 
&\scalebox{0.78}{7} &\scalebox{0.78}{17420} &\scalebox{0.78}{1-hour} &\scalebox{0.78}{Electricity}  \\
\midrule
{{\scalebox{0.95}{ETTm1}}} 
&\scalebox{0.78}{7} &\scalebox{0.78}{69680} &\scalebox{0.78}{15-min} &\scalebox{0.78}{Electricity}  \\
\midrule
{{\scalebox{0.95}{ETTm2}}} 
&\scalebox{0.78}{7} &\scalebox{0.78}{69680} &\scalebox{0.78}{15-min} &\scalebox{0.78}{Electricity}  \\
\midrule
{{\scalebox{0.95}{Electricity}}} 
&\scalebox{0.78}{321} &\scalebox{0.78}{26304} &\scalebox{0.78}{1-hour} &\scalebox{0.78}{Electricity}  \\
\midrule
{{\scalebox{0.95}{Traffic}}} 
&\scalebox{0.78}{862} &\scalebox{0.78}{17544} &\scalebox{0.78}{1-hour} &\scalebox{0.78}{Transportation}  \\
\midrule
{{\scalebox{0.95}{Weather}}} 
&\scalebox{0.78}{21} &\scalebox{0.78}{52696} &\scalebox{0.78}{10-min} &\scalebox{0.78}{Weather}  \\
\midrule
{{\scalebox{0.95}{Solar}}} 
&\scalebox{0.78}{137} &\scalebox{0.78}{52560} &\scalebox{0.78}{10-min} &\scalebox{0.78}{Energy}  \\
    
    \bottomrule
  \end{tabular}
    \end{small}
  \end{threeparttable}
\end{table}

\subsection{Baseline Models}
\textbf{Autoformer \cite{wu2021autoformer}:} Autoformer introduces time series decomposition into the Transformer architecture, and designs an Auto-Correlation mechanism to take the place of self-attention to improve computation efficiency and information utilization.

\textbf{FEDformer \cite{zhou2022fedformer}:} FEDformer designs a Frequency Enhanced Block with 2 versions based on DFT (Discrete Fourier Transform) and DWT (Discrete Wavelet Transform) respectively to capture features in frequency domain. It also utilizes mix-of-experts decomposition to deal with distribution shifting. In our experiments, we employs the Fourier version for FEDformer.

\textbf{iTransformer \cite{liu2023itransformer}:} iTransformer reverses the roles of the attention mechanism and the feed-forward network by embedding the time points of each individual series into variate tokens, which are then utilized by the attention mechanism to capture multivariate correlations.

\textbf{Crossformer \cite{zhang2022crossformer}:} Crossformer introduces the Dimension-Segment-Wise (DSW) embedding and Two-Stage Attention (TSA) techniques to enhance the capturing of cross-dimension dependencies in multi-variate time series forecasting.

\textbf{PatchTST \cite{nie2022time}:} PatchTST proposes an effective design of Transformer-based models for time series forecasting tasks by introducing two key components: patching and channel-independent structure. We take the default setting that Patch\_len=16 and Stride=8.

\textbf{TimesNet \cite{wu2022timesnet}:} TimesNet transforms 1D time series into 2D space according to the main periods detected from the topk frequency spectrum amplitude by FFT. It utilizes a parameter-efficient inception block to capture the intra-period and inter-period dependencies at the same time in the 2D space. 

\textbf{SCINet \cite{liu2022scinet}:} SCINet employs a recursive downsample-convolve-interact architecture, where multiple convolutional filters are utilized in each layer to extract distinct and valuable temporal features from downsampled sub-sequences or features.

\textbf{DLinear \cite{zeng2023transformers}:} DLinear proposes a simple one layer MLP structure for forecasting. Specifically, it employs decomposition and applies 2 MLPs to the trend and seasonal parts, respectively.

\textbf{FiLM \cite{zhou2022film}:} FilM  employs the Legendre Polynomials projection technique to approximate historical information, utilizes FFT for frequency feature extraction by Frequency Enhanced Layer.

\textbf{FreTS \cite{yi2023frequency}:} FreTS designs a frequency domain MLP structure based on the operation of complex number, which is applied in the frequency domain. It contains 2 stages on both inter-series and intra-series scales, to better capture channel-wise and time-wise dependencies in the frequency domain simultaneously.

\textbf{FITS \cite{xu2024fits}:} FITS proposes a lightweight model that utilizes complex-valued linear layer to capture frequency domain features. It also contains a low-pass filter to deal with the high frequency noise in the time series.

\textbf{StemGNN \cite{cao2020spectral}:} StemGNN effectively captures inter-series correlations and temporal dependencies simultaneously by leveraging the frequency domain. This is achieved through an end-to-end framework that combines the Graph Fourier Transform (GFT) to model inter-series correlations and the Discrete Fourier Transform (DFT) to model temporal dependencies.

We implement \textbf{Autoformer}, \textbf{FEDformer}, \textbf{iTransformer}, \textbf{Crossformer}, \textbf{PatchTST}, \textbf{TimesNet}, \textbf{DLinear}, \textbf{FiLM} based on the Time-Series-Library repository: \href{https://github.com/thuml/Time-Series-Library}{https://github.com/thuml/Time-Series-Library}. 

For the rest models, we implement them based on their official code repository.

\textbf{SCINet}: \href{https://github.com/cure-lab/SCINet}{https://github.com/cure-lab/SCINet}. 

\textbf{FreTS}: \href{https://github.com/aikunyi/FreTS}{https://github.com/aikunyi/FreTS}. 

\textbf{FITS}: \href{https://github.com/VEWOXIC/FITS}{https://github.com/VEWOXIC/FITS}. 

\textbf{StemGNN}:\href{https://github.com/microsoft/StemGNN/}{https://github.com/microsoft/StemGNN/}.

\subsection{Implementation Details}
We implement our model and all baseline models in PyTorch \cite{paszke2019pytorch}, and train the models on NVIDIA A100-SXM 80GB GPUs. We choose ADAM \cite{kingma2014adam} as optimizer with a initial learning rate of $1e^{-4}$. The batch size is set differently to distinct datasets according to the number of variate. Specifically, for Traffic and Electricity dataset, we set the batch size as 32. For Solar dataset, we set the batch size as 128, and for the rest datasets, we set the batch size as 256. We employ an early stopping counter to stop the training after 3 epochs if there is no loss degradation on the valid set. By default, the F-Block contains 2 layers of complex-valued Transformer Encoder with the dimension of 512. We keep the same hyper-parameters setting for all models and all experiments.

We reproduce Autoformer, FEDformer, iTransformer, Crossformer, PatchTST, TimesNet, DLinear, FiLM based on the TimesNet \cite{wu2022timesnet} Repository, and SCINet, FreTS, FITS, StemGNN based on their official code repository. 

\subsection{Robustness of \model\ performance}
To test the robustness of \model\, we report the standard deviation of \model\ performance by repeating each experiment for 5 times under distinct random seeds. The results are listed in Table \ref{std}, through which we can learn that the performance of \model\ is stable.

\begin{table}[htbp]
  \vskip 0.05in
  \caption{Multivariate long-term forecasting results of \model\ with 5 different random seeds.}
  \label{std}
  \centering
  \begin{threeparttable}
  \begin{small}
  \renewcommand{\multirowsetup}{\centering}
  \setlength{\tabcolsep}{0.8pt}
  \begin{tabular}{c|c|cc|cc|cc|cc|cc|cc}
    \toprule
    \multicolumn{2}{c}{Dataset} & 
    \multicolumn{2}{c}{\rotatebox{0}{\scalebox{0.68}{{ETTh1}}}} & \multicolumn{2}{c}{\rotatebox{0}{\scalebox{0.68}{{ETTh2}}}} & \multicolumn{2}{c}{\rotatebox{0}{\scalebox{0.68}{{ETTm1}}}} & \multicolumn{2}{c}{\rotatebox{0}{\scalebox{0.68}{{ETTm2}}}}\\
    
    \cmidrule(lr){3-4} \cmidrule(lr){5-6}\cmidrule(lr){7-8} \cmidrule(lr){9-10}
    \multicolumn{2}{c}{} & \scalebox{0.68}{MSE} & \scalebox{0.68}{MAE} & \scalebox{0.68}{MSE} & \scalebox{0.68}{MAE} & \scalebox{0.68}{MSE} & \scalebox{0.68}{MAE} & \scalebox{0.68}{MSE} & \scalebox{0.68}{MAE}   \\ \midrule

    \multirow{5}{*}{\rotatebox{90}{\scalebox{0.95}{ATFNet}}}
    &  \scalebox{0.78}{96} &\scalebox{0.78}{0.409 $\pm$ 0.002} &\scalebox{0.78}{0.442 $\pm$ 0.000} &\scalebox{0.78}{0.172 $\pm$ 0.001} &\scalebox{0.78}{0.283 $\pm$ 0.000} &\scalebox{0.78}{0.324 $\pm$ 0.003} &\scalebox{0.78}{0.376 $\pm$ 0.001} &\scalebox{0.78}{0.114 $\pm$ 0.001} &\scalebox{0.78}{0.227 $\pm$ 0.001} \\
&  \scalebox{0.78}{192} &\scalebox{0.78}{0.459 $\pm$ 0.005} &\scalebox{0.78}{0.481 $\pm$ 0.002} &\scalebox{0.78}{0.212 $\pm$ 0.003} &\scalebox{0.78}{0.318 $\pm$ 0.003} &\scalebox{0.78}{0.370 $\pm$ 0.005} &\scalebox{0.78}{0.411 $\pm$ 0.002} &\scalebox{0.78}{0.143 $\pm$ 0.002} &\scalebox{0.78}{0.255 $\pm$ 0.002} \\
&  \scalebox{0.78}{336} &\scalebox{0.78}{0.509 $\pm$ 0.007} &\scalebox{0.78}{0.517 $\pm$ 0.004} &\scalebox{0.78}{0.242 $\pm$ 0.004} &\scalebox{0.78}{0.344 $\pm$ 0.005} &\scalebox{0.78}{0.411 $\pm$ 0.005} &\scalebox{0.78}{0.442 $\pm$ 0.002} &\scalebox{0.78}{0.176 $\pm$ 0.002} &\scalebox{0.78}{0.284 $\pm$ 0.002} \\
&  \scalebox{0.78}{720} &\scalebox{0.78}{0.610 $\pm$ 0.005} &\scalebox{0.78}{0.585 $\pm$ 0.001} &\scalebox{0.78}{0.308 $\pm$ 0.008} &\scalebox{0.78}{0.409 $\pm$ 0.006} &\scalebox{0.78}{0.459 $\pm$ 0.004} &\scalebox{0.78}{0.481 $\pm$ 0.003} &\scalebox{0.78}{0.221 $\pm$ 0.002} &\scalebox{0.78}{0.320 $\pm$ 0.002} \\

    \midrule
    \midrule
    \multicolumn{2}{c}{Dataset} & 
    \multicolumn{2}{c}{\rotatebox{0}{\scalebox{0.68}{{Traffic}}}} & \multicolumn{2}{c}{\rotatebox{0}{\scalebox{0.68}{{Electricity}}}} & \multicolumn{2}{c}{\rotatebox{0}{\scalebox{0.68}{{Weather}}}} & \multicolumn{2}{c}{\rotatebox{0}{\scalebox{0.68}{{Solar}}}}\\
    
    \cmidrule(lr){3-4} \cmidrule(lr){5-6}\cmidrule(lr){7-8} \cmidrule(lr){9-10}
    \multicolumn{2}{c}{} & \scalebox{0.68}{MSE} & \scalebox{0.68}{MAE} & \scalebox{0.68}{MSE} & \scalebox{0.68}{MAE} & \scalebox{0.68}{MSE} & \scalebox{0.68}{MAE} & \scalebox{0.68}{MSE} & \scalebox{0.68}{MAE}   \\ \midrule
\multirow{5}{*}{\rotatebox{90}{\scalebox{0.95}{ATFNet}}}
    &  \scalebox{0.78}{96} &\scalebox{0.78}{0.376 $\pm$ 0.001} &\scalebox{0.78}{0.264 $\pm$ 0.001} &\scalebox{0.78}{0.131 $\pm$ 0.001} &\scalebox{0.78}{0.228 $\pm$ 0.001} &\scalebox{0.78}{0.158 $\pm$ 0.002} &\scalebox{0.78}{0.208 $\pm$ 0.002} &\scalebox{0.78}{0.182 $\pm$ 0.001} &\scalebox{0.78}{0.242 $\pm$ 0.002} \\
&  \scalebox{0.78}{192} &\scalebox{0.78}{0.393 $\pm$ 0.001} &\scalebox{0.78}{0.271 $\pm$ 0.001} &\scalebox{0.78}{0.148 $\pm$ 0.000} &\scalebox{0.78}{0.244 $\pm$ 0.000} &\scalebox{0.78}{0.199 $\pm$ 0.001} &\scalebox{0.78}{0.248 $\pm$ 0.002} &\scalebox{0.78}{0.198 $\pm$ 0.001} &\scalebox{0.78}{0.260 $\pm$ 0.002} \\
&  \scalebox{0.78}{336} &\scalebox{0.78}{0.407 $\pm$ 0.000} &\scalebox{0.78}{0.278 $\pm$ 0.000} &\scalebox{0.78}{0.164 $\pm$ 0.000} &\scalebox{0.78}{0.262 $\pm$ 0.000} &\scalebox{0.78}{0.248 $\pm$ 0.001} &\scalebox{0.78}{0.287 $\pm$ 0.002} &\scalebox{0.78}{0.204 $\pm$ 0.001} &\scalebox{0.78}{0.266 $\pm$ 0.002} \\
&  \scalebox{0.78}{720} &\scalebox{0.78}{0.438 $\pm$ 0.001} &\scalebox{0.78}{0.298 $\pm$ 0.001} &\scalebox{0.78}{0.199 $\pm$ 0.002} &\scalebox{0.78}{0.295 $\pm$ 0.001} &\scalebox{0.78}{0.312 $\pm$ 0.001} &\scalebox{0.78}{0.336 $\pm$ 0.001} &\scalebox{0.78}{0.209 $\pm$ 0.001} &\scalebox{0.78}{0.278 $\pm$ 0.003} \\

    \bottomrule
  \end{tabular}
    \end{small}
  \end{threeparttable}
\end{table}

\subsection{Model efficiency comparison}
We assess the computational and memory requirements of \model\ in comparison to other baseline models. This evaluation is performed on the ETTh1 dataset using a input-96-predict-96 configuration. We maintain consistent configurations across all models, aligning them with the main experiment for multi-variate long-term forecasting. The results are presented in Table \ref{efficiency}.
\begin{table}[ht]
\centering
\begin{threeparttable}
\begin{small}
  \renewcommand{\multirowsetup}{\centering}
  \setlength{\tabcolsep}{0.8pt}
\caption{Comparison of computational and memory costs}
\label{efficiency}
\begin{tabular}{c|c|c|c|c|c|c|c|c|c|c|c|c|c}
\toprule
& \scalebox{0.8}{\model} & \scalebox{0.8}{Autoformer} & \scalebox{0.8}{Crossformer} & \scalebox{0.8}{DLinear} & \scalebox{0.8}{FEDformer} & \scalebox{0.8}{FITS} & \scalebox{0.8}{FiLM} & \scalebox{0.8}{FreTS} & \scalebox{0.8}{PatchTST} & \scalebox{0.8}{SCINet} & \scalebox{0.8}{StemGNN} & \scalebox{0.8}{TimesNet} & \scalebox{0.8}{iTransformer} \\
\midrule
\scalebox{0.8}{Memory (GB)} & \scalebox{0.78}{0.974} & \scalebox{0.78}{0.884} & \scalebox{0.78}{0.402} & \scalebox{0.78}{0.001} & \scalebox{0.78}{1.409} & \scalebox{0.78}{0.001} & \scalebox{0.78}{0.423} & \scalebox{0.78}{0.146} & \scalebox{0.78}{0.320} & \scalebox{0.78}{0.004} & \scalebox{0.78}{1.305} & \scalebox{0.78}{0.054} & \scalebox{0.78}{0.123} \\
\scalebox{0.8}{Training Speed (ms/iter)} & \scalebox{0.78}{48.0} & \scalebox{0.78}{29.0} & \scalebox{0.78}{57.0} & \scalebox{0.78}{7.0} & \scalebox{0.78}{176.0} & \scalebox{0.78}{8.0} & \scalebox{0.78}{65.0} & \scalebox{0.78}{11.0} & \scalebox{0.78}{15.0} & \scalebox{0.78}{50.0} & \scalebox{0.78}{21.0} & \scalebox{0.78}{30.0} & \scalebox{0.78}{15.0} \\
\bottomrule
\end{tabular}
\end{small}
\end{threeparttable}
\end{table}

\subsection{RevIN\cite{kim2021reversible} in F-Block}
As we discuss in section \ref{F-Block}, we apply RevIN in F-Block to address distribution drift problem in frequency domain. In this experiment we test the effectiveness of RevIN in F-Block. Specifically, we train a new version of \model\ without complex-valued RevIN in F-Block. To maintain fairness comparison, we apply origin RevIN on the whole model rather than only keep RevIN in T-Block. The results are listed in Table \ref{RevIN}.

\begin{table}[ht]
\centering
\begin{threeparttable}
\begin{small}
\renewcommand{\multirowsetup}{\centering}
\setlength{\tabcolsep}{0.8pt}
\caption{Multi-variate long-term forecasting results of \model\ with and without RevIN in F-Block. \model\ (w/o) represents \model\ without RevIN in F-Block.}
\label{RevIN}
\label{efficiency}
\begin{tabular}{c|c|cc|cc|cc|cc|cc|cc|cc|cc}
\toprule
\multicolumn{2}{c}{Datasets} & \multicolumn{2}{c}{\scalebox{0.8}{ETTh1}} & \multicolumn{2}{c}{\scalebox{0.8}{ETTh2}} & \multicolumn{2}{c}{\scalebox{0.8}{ETTm1}} & \multicolumn{2}{c}{\scalebox{0.8}{ETTm2}} & \multicolumn{2}{c}{\scalebox{0.8}{Traffic}} & \multicolumn{2}{c}{\scalebox{0.8}{Electricity}} & \multicolumn{2}{c}{\scalebox{0.8}{Weather}} & \multicolumn{2}{c}{\scalebox{0.8}{Solar}} \\
\cmidrule(r){3-4} \cmidrule(lr){5-6} \cmidrule(lr){7-8} \cmidrule(lr){9-10} \cmidrule(lr){11-12} \cmidrule(lr){13-14} \cmidrule(lr){15-16} \cmidrule(l){17-18} 
\multicolumn{2}{c}{}& \scalebox{0.78}{MSE} & \scalebox{0.78}{MAE} & \scalebox{0.78}{MSE} & \scalebox{0.78}{MAE} & \scalebox{0.78}{MSE} & \scalebox{0.78}{MAE} & \scalebox{0.78}{MSE} & \scalebox{0.78}{MAE} & \scalebox{0.78}{MSE} & \scalebox{0.78}{MAE} & \scalebox{0.78}{MSE} & \scalebox{0.78}{MAE} & \scalebox{0.78}{MSE} & \scalebox{0.78}{MAE} & \scalebox{0.78}{MSE} & \scalebox{0.78}{MAE} \\
\midrule
\multirow{4}{*}{\rotatebox{90}{\scalebox{0.78}{ATFNet}}}  
& \scalebox{0.78}{96} & \scalebox{0.78}{0.405} & \scalebox{0.78}{0.442} & \scalebox{0.78}{0.171} & \scalebox{0.78}{0.282} & \scalebox{0.78}{0.327} & \scalebox{0.78}{0.375} & \scalebox{0.78}{0.114} & \scalebox{0.78}{0.227} & \scalebox{0.78}{0.376} & \scalebox{0.78}{0.264} & \scalebox{0.78}{0.131} & \scalebox{0.78}{0.227} & \scalebox{0.78}{0.156} & \scalebox{0.78}{0.206} & \scalebox{0.78}{0.180} & \scalebox{0.78}{0.241} \\
& \scalebox{0.78}{192} & \scalebox{0.78}{0.467} & \scalebox{0.78}{0.482} & \scalebox{0.78}{0.213} & \scalebox{0.78}{0.315} & \scalebox{0.78}{0.370} & \scalebox{0.78}{0.412} & \scalebox{0.78}{0.141} & \scalebox{0.78}{0.254} & \scalebox{0.78}{0.394} & \scalebox{0.78}{0.273} & \scalebox{0.78}{0.148} & \scalebox{0.78}{0.244} & \scalebox{0.78}{0.199} & \scalebox{0.78}{0.246} & \scalebox{0.78}{0.200} & \scalebox{0.78}{0.264} \\
& \scalebox{0.78}{336} & \scalebox{0.78}{0.514} & \scalebox{0.78}{0.521} & \scalebox{0.78}{0.236} & \scalebox{0.78}{0.344} & \scalebox{0.78}{0.402} & \scalebox{0.78}{0.438} & \scalebox{0.78}{0.172} & \scalebox{0.78}{0.281} & \scalebox{0.78}{0.406} & \scalebox{0.78}{0.278} & \scalebox{0.78}{0.164} & \scalebox{0.78}{0.262} & \scalebox{0.78}{0.249} & \scalebox{0.78}{0.286} & \scalebox{0.78}{0.202} & \scalebox{0.78}{0.266} \\
& \scalebox{0.78}{720} & \scalebox{0.78}{0.614} & \scalebox{0.78}{0.589} & \scalebox{0.78}{0.297} & \scalebox{0.78}{0.399} & \scalebox{0.78}{0.453} & \scalebox{0.78}{0.476} & \scalebox{0.78}{0.219} & \scalebox{0.78}{0.317} & \scalebox{0.78}{0.437} & \scalebox{0.78}{0.299} & \scalebox{0.78}{0.200} & \scalebox{0.78}{0.295} & \scalebox{0.78}{0.311} & \scalebox{0.78}{0.336} & \scalebox{0.78}{0.208} & \scalebox{0.78}{0.273} \\
\midrule
\multirow{4}{*}{\rotatebox{90}{\scalebox{0.78}{ATFNet(w/o)}}}  
& \scalebox{0.78}{96} & \scalebox{0.78}{0.429} & \scalebox{0.78}{0.456} & \scalebox{0.78}{0.171} & \scalebox{0.78}{0.284} & \scalebox{0.78}{0.329} & \scalebox{0.78}{0.380} & \scalebox{0.78}{0.115} & \scalebox{0.78}{0.229} & \scalebox{0.78}{0.405} & \scalebox{0.78}{0.297} & \scalebox{0.78}{0.145} & \scalebox{0.78}{0.249} & \scalebox{0.78}{0.152} & \scalebox{0.78}{0.202} & \scalebox{0.78}{0.189} & \scalebox{0.78}{0.255} \\
& \scalebox{0.78}{192} & \scalebox{0.78}{0.500} & \scalebox{0.78}{0.499} & \scalebox{0.78}{0.218} & \scalebox{0.78}{0.320} & \scalebox{0.78}{0.377} & \scalebox{0.78}{0.412} & \scalebox{0.78}{0.146} & \scalebox{0.78}{0.258} & \scalebox{0.78}{0.418} & \scalebox{0.78}{0.301} & \scalebox{0.78}{0.161} & \scalebox{0.78}{0.263} & \scalebox{0.78}{0.197} & \scalebox{0.78}{0.243} & \scalebox{0.78}{0.212} & \scalebox{0.78}{0.271} \\
& \scalebox{0.78}{336} & \scalebox{0.78}{0.552} & \scalebox{0.78}{0.532} & \scalebox{0.78}{0.248} & \scalebox{0.78}{0.345} & \scalebox{0.78}{0.416} & \scalebox{0.78}{0.441} & \scalebox{0.78}{0.175} & \scalebox{0.78}{0.287} & \scalebox{0.78}{0.432} & \scalebox{0.78}{0.308} & \scalebox{0.78}{0.176} & \scalebox{0.78}{0.278} & \scalebox{0.78}{0.248} & \scalebox{0.78}{0.284} & \scalebox{0.78}{0.222} & \scalebox{0.78}{0.281} \\
& \scalebox{0.78}{720} & \scalebox{0.78}{0.725} & \scalebox{0.78}{0.642} & \scalebox{0.78}{0.324} & \scalebox{0.78}{0.401} & \scalebox{0.78}{0.474} & \scalebox{0.78}{0.485} & \scalebox{0.78}{0.224} & \scalebox{0.78}{0.323} & \scalebox{0.78}{0.464} & \scalebox{0.78}{0.325} & \scalebox{0.78}{0.211} & \scalebox{0.78}{0.303} & \scalebox{0.78}{0.321} & \scalebox{0.78}{0.336} & \scalebox{0.78}{0.220} & \scalebox{0.78}{0.279} \\

\bottomrule
\end{tabular}
\end{small}
\end{threeparttable}
\end{table}

\subsection{Forecasting Results Visualization}
\begin{figure*}[htbp]
\centering
\includegraphics[width=6.7in]{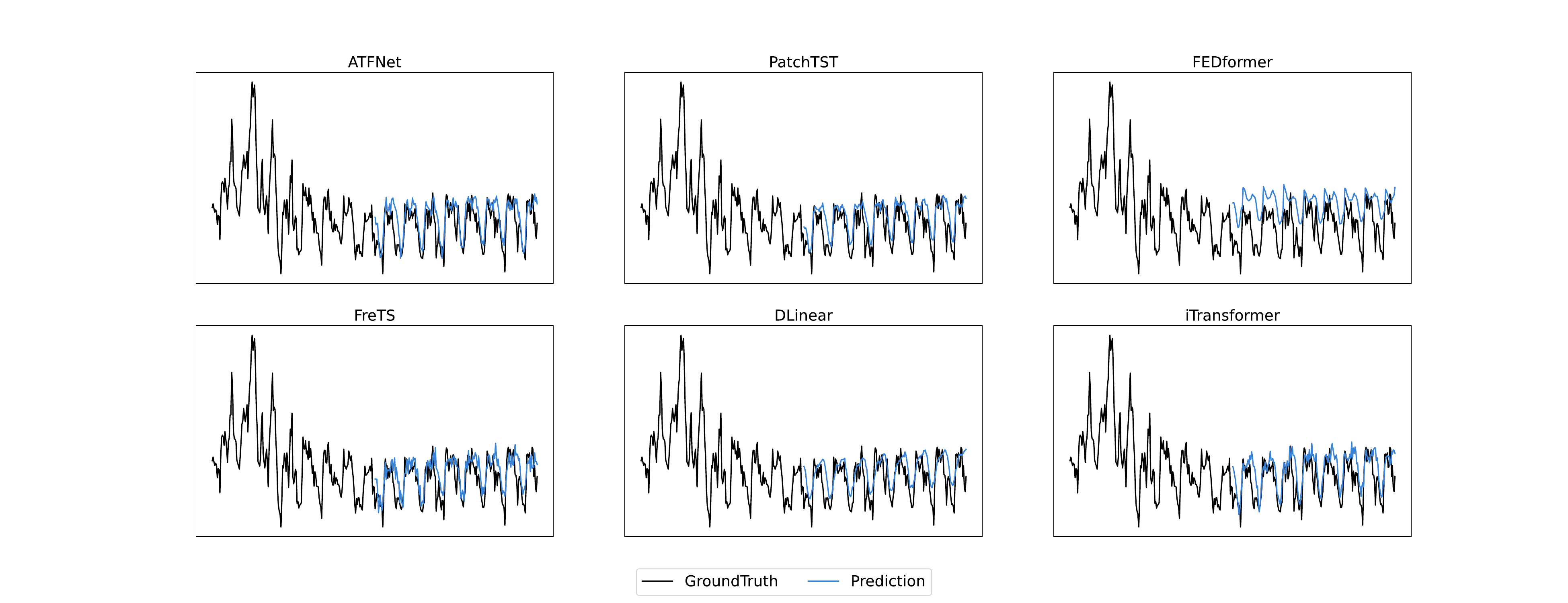}
\centering
\caption{Visualization of the input-192-predict-192 forecasting results on ETTh1 dataset. }
\label{ETTh1}
\end{figure*}

\begin{figure*}[htbp]
\centering
\includegraphics[width=6.7in]{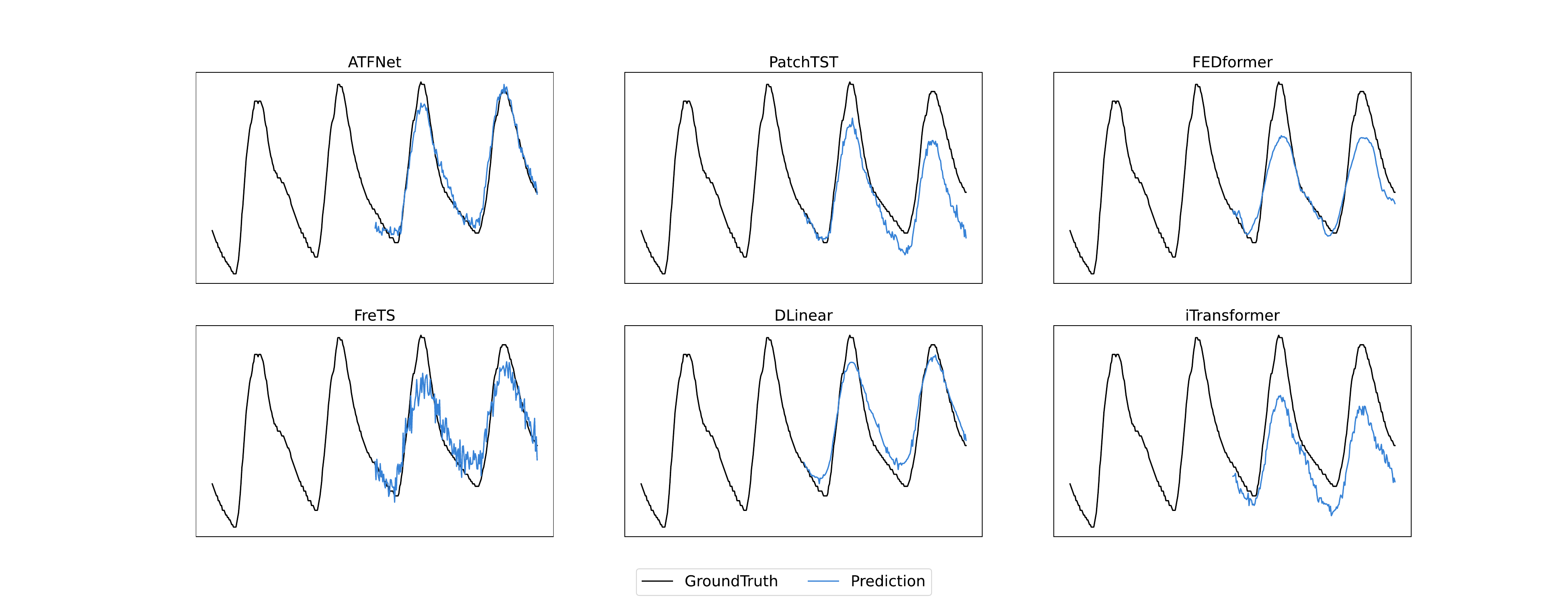}
\centering
\caption{Visualization of the input-192-predict-192 forecasting results on ETTm2 dataset. }
\label{ETTm2}
\end{figure*}

\begin{figure*}[htbp]
\centering
\includegraphics[width=6.7in]{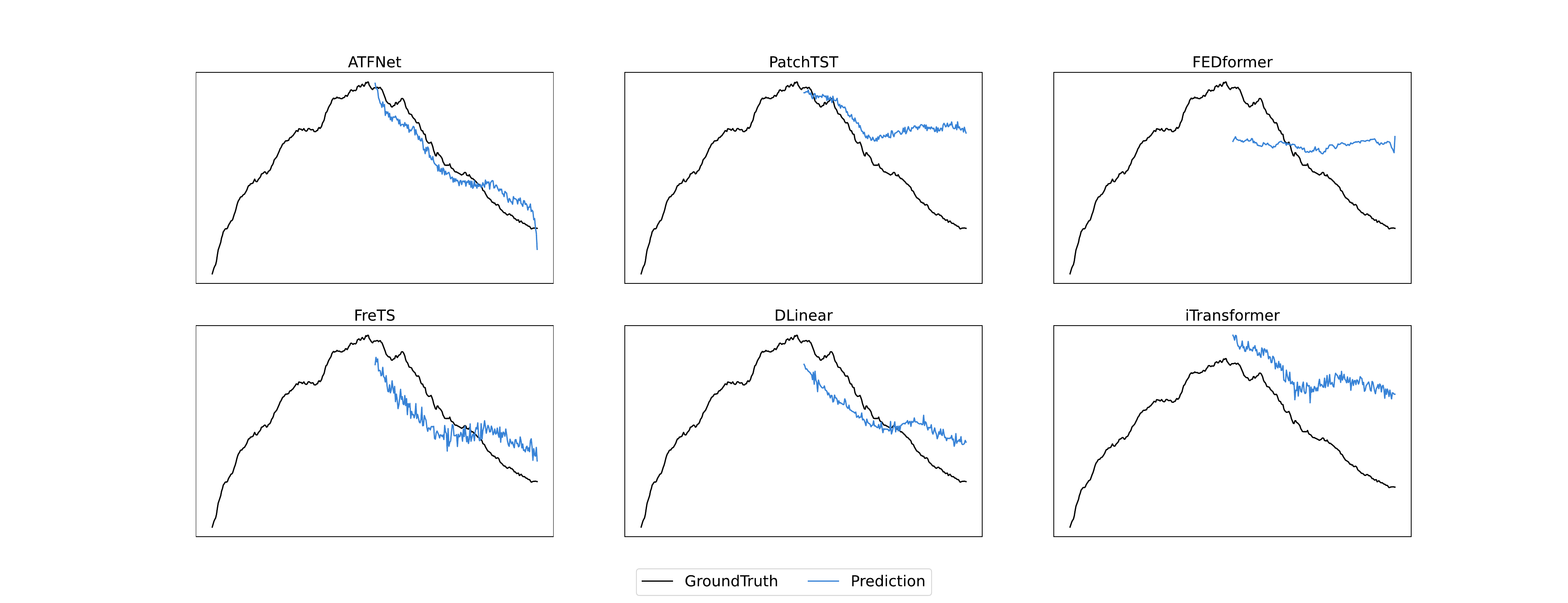}
\centering
\caption{Visualization of the input-192-predict-192 forecasting results on Weather dataset. }
\label{weather_}
\end{figure*}

In order to facilitate a clear comparison between different models, we present supplementary prediction demonstrations for three representative datasets in Figures \ref{ETTh1}-\ref{weather_}.

\subsection{Full Results}
\label{full_result}
\begin{table*}[h]
  \caption{Full Results of Multi-variate long-term time series forecasting.}
  \label{full_multi}
  \vskip 0.05in
  \centering
  \begin{threeparttable}
  \begin{small}
  \renewcommand{\multirowsetup}{\centering}
  \setlength{\tabcolsep}{0.8pt}
\begin{tabular}{c|c|cc|cc|cc|cc|cc|cc|cc|cc|cc|cc|cc|cc|cc}
    \toprule
\multicolumn{2}{c}{Models} &  
\multicolumn{2}{c}{\rotatebox{0}{\scalebox{0.8}{{\model}}}} &
\multicolumn{2}{c}{\rotatebox{0}{\scalebox{0.8}{{Autoformer}}}} &
\multicolumn{2}{c}{\rotatebox{0}{\scalebox{0.8}{{FEDformer}}}} &
\multicolumn{2}{c}{\rotatebox{0}{\scalebox{0.8}{{iTransformer}}}} &
\multicolumn{2}{c}{\rotatebox{0}{\scalebox{0.8}{{Crossformer}}}} &
\multicolumn{2}{c}{\rotatebox{0}{\scalebox{0.8}{{PatchTST}}}} &
\multicolumn{2}{c}{\rotatebox{0}{\scalebox{0.8}{{TimesNet}}}} &
\multicolumn{2}{c}{\rotatebox{0}{\scalebox{0.8}{{SCINet}}}} &
\multicolumn{2}{c}{\rotatebox{0}{\scalebox{0.8}{{DLinear}}}} &
\multicolumn{2}{c}{\rotatebox{0}{\scalebox{0.8}{{FiLM}}}} &
\multicolumn{2}{c}{\rotatebox{0}{\scalebox{0.8}{{FreTS}}}} &
\multicolumn{2}{c}{\rotatebox{0}{\scalebox{0.8}{{FITS}}}} &
\multicolumn{2}{c}{\rotatebox{0}{\scalebox{0.8}{{StemGNN}}}} \\
    
    \cmidrule(lr){3-4}\cmidrule(lr){5-6}\cmidrule(lr){7-8}\cmidrule(lr){9-10}\cmidrule(lr){11-12}\cmidrule(lr){13-14}\cmidrule(lr){15-16}
    \cmidrule(lr){17-18}\cmidrule(lr){19-20}\cmidrule(lr){21-22}\cmidrule(lr){23-24}\cmidrule(lr){25-26}\cmidrule(lr){27-28}
    \multicolumn{2}{c}{} & \scalebox{0.68}{MSE} & \scalebox{0.68}{MAE} & \scalebox{0.68}{MSE} & \scalebox{0.68}{MAE} & \scalebox{0.68}{MSE} &  \scalebox{0.68}{MAE} & \scalebox{0.68}{MSE} & \scalebox{0.68}{MAE} & \scalebox{0.68}{MSE} & \scalebox{0.68}{MAE} & \scalebox{0.68}{MSE} & \scalebox{0.68}{MAE} & \scalebox{0.68}{MSE} & \scalebox{0.68}{MAE} & \scalebox{0.68}{MSE} & \scalebox{0.68}{MAE} & \scalebox{0.68}{MSE} & \scalebox{0.68}{MAE} & \scalebox{0.68}{MSE} & \scalebox{0.68}{MAE} & \scalebox{0.68}{MSE} & \scalebox{0.68}{MAE} & \scalebox{0.68}{MSE} & \scalebox{0.68}{MAE} & \scalebox{0.68}{MSE} & \scalebox{0.68}{MAE}  \\
\bottomrule
  \end{tabular}
    \end{small}
  \end{threeparttable}
\end{table*}

\begin{table*}[htbp]
  \caption{Full Results of Uni-variate long-term time series forecasting.}
  \label{full_uni}
  \vskip 0.05in
  \centering
  \begin{threeparttable}
  \begin{small}
  \renewcommand{\multirowsetup}{\centering}
  \setlength{\tabcolsep}{0.8pt}
  \begin{tabular}{c|c|cc|cc|cc|cc|cc|cc|cc|cc|cc|ccc}
    \toprule
\multicolumn{2}{c}{Models} &  
\multicolumn{2}{c}{\rotatebox{0}{\scalebox{0.8}{{\model}}}} &
\multicolumn{2}{c}{\rotatebox{0}{\scalebox{0.8}{{Autoformer}}}} &
\multicolumn{2}{c}{\rotatebox{0}{\scalebox{0.8}{{FEDformer}}}} &
\multicolumn{2}{c}{\rotatebox{0}{\scalebox{0.8}{{PatchTST}}}} &
\multicolumn{2}{c}{\rotatebox{0}{\scalebox{0.8}{{TimesNet}}}} &
\multicolumn{2}{c}{\rotatebox{0}{\scalebox{0.8}{{SCINet}}}} &
\multicolumn{2}{c}{\rotatebox{0}{\scalebox{0.8}{{DLinear}}}} &
\multicolumn{2}{c}{\rotatebox{0}{\scalebox{0.8}{{FiLM}}}} &
\multicolumn{2}{c}{\rotatebox{0}{\scalebox{0.8}{{FreTS}}}} &
\multicolumn{2}{c}{\rotatebox{0}{\scalebox{0.8}{{FITS}}}} & \\
    
    \cmidrule(lr){3-4}\cmidrule(lr){5-6}\cmidrule(lr){7-8}\cmidrule(lr){9-10}\cmidrule(lr){11-12}\cmidrule(lr){13-14}\cmidrule(lr){15-16}\cmidrule(lr){17-18}\cmidrule(lr){19-20}\cmidrule(lr){21-22}
    \multicolumn{2}{c}{} & \scalebox{0.68}{MSE} & \scalebox{0.68}{MAE} & \scalebox{0.68}{MSE} & \scalebox{0.68}{MAE} & \scalebox{0.68}{MSE} &  \scalebox{0.68}{MAE} & \scalebox{0.68}{MSE} & \scalebox{0.68}{MAE} & \scalebox{0.68}{MSE} & \scalebox{0.68}{MAE} & \scalebox{0.68}{MSE} & \scalebox{0.68}{MAE} & \scalebox{0.68}{MSE} & \scalebox{0.68}{MAE} & \scalebox{0.68}{MSE} & \scalebox{0.68}{MAE} & \scalebox{0.68}{MSE} & \scalebox{0.68}{MAE} & \scalebox{0.68}{MSE} & \scalebox{0.68}{MAE}   \\
\bottomrule
  \end{tabular}
    \end{small}
  \end{threeparttable}
\end{table*}

\begin{table}[htbp]
  \caption{Full Results of Ablation Study.}
  \label{full_ablation}
  \vskip 0.05in
  \centering
  \begin{threeparttable}
  \begin{small}
  \renewcommand{\multirowsetup}{\centering}
  \setlength{\tabcolsep}{0.8pt}
  \begin{tabular}{c|c|cc|cc|cc|cc|cc|cc|cc|cc|cc|cc|cc|cc|cc}
    \toprule
\multicolumn{2}{c}{Models} &  
\multicolumn{2}{c}{\rotatebox{0}{\scalebox{0.8}{{ATFNet}}}} &
\multicolumn{2}{c}{\rotatebox{0}{\scalebox{0.8}{{F-Block}}}} &
\multicolumn{2}{c}{\rotatebox{0}{\scalebox{0.8}{{T-Block}}}} &
\multicolumn{2}{c}{\rotatebox{0}{\scalebox{0.8}{{ATFNet v1}}}} &
\multicolumn{2}{c}{\rotatebox{0}{\scalebox{0.8}{{ATFNet v2}}}} &
\multicolumn{2}{c}{\rotatebox{0}{\scalebox{0.8}{{ATFNet v3}}}}  \\
    
    \cmidrule(lr){3-4}\cmidrule(lr){5-6}\cmidrule(lr){7-8}\cmidrule(lr){9-10}\cmidrule(lr){11-12}\cmidrule(lr){13-14}
    \multicolumn{2}{c}{} & \scalebox{0.68}{MSE} & \scalebox{0.68}{MAE} & \scalebox{0.68}{MSE} & \scalebox{0.68}{MAE} & \scalebox{0.68}{MSE} &  \scalebox{0.68}{MAE} & \scalebox{0.68}{MSE} & \scalebox{0.68}{MAE} & \scalebox{0.68}{MSE} & \scalebox{0.68}{MAE} & \scalebox{0.68}{MSE} & \scalebox{0.68}{MAE}   \\ \midrule

\multirow{5}{*}{\rotatebox{90}{\scalebox{0.95}{ETTh1}}} 
&  \scalebox{0.78}{96} &\boldres{\scalebox{0.78}{0.405}} &\scalebox{0.78}{0.442} &\secondres{\scalebox{0.78}{0.413}} &\scalebox{0.78}{0.445} &\scalebox{0.78}{0.419} &\scalebox{0.78}{0.448} &\scalebox{0.78}{0.414} &\boldres{\scalebox{0.78}{0.441}} &\scalebox{0.78}{0.414} &\secondres{\scalebox{0.78}{0.441}} &\scalebox{0.78}{0.417} &\scalebox{0.78}{0.444}  \\
&  \scalebox{0.78}{192} &\boldres{\scalebox{0.78}{0.467}} &\boldres{\scalebox{0.78}{0.482}} &\secondres{\scalebox{0.78}{0.469}} &\secondres{\scalebox{0.78}{0.483}} &\scalebox{0.78}{0.482} &\scalebox{0.78}{0.488} &\scalebox{0.78}{0.471} &\scalebox{0.78}{0.485} &\scalebox{0.78}{0.470} &\scalebox{0.78}{0.487} &\scalebox{0.78}{0.475} &\scalebox{0.78}{0.488}  \\
\bottomrule
  \end{tabular}
    \end{small}
  \end{threeparttable}
\end{table}

\begin{figure}[htbp]
\centering
\includegraphics[width=6.5in]{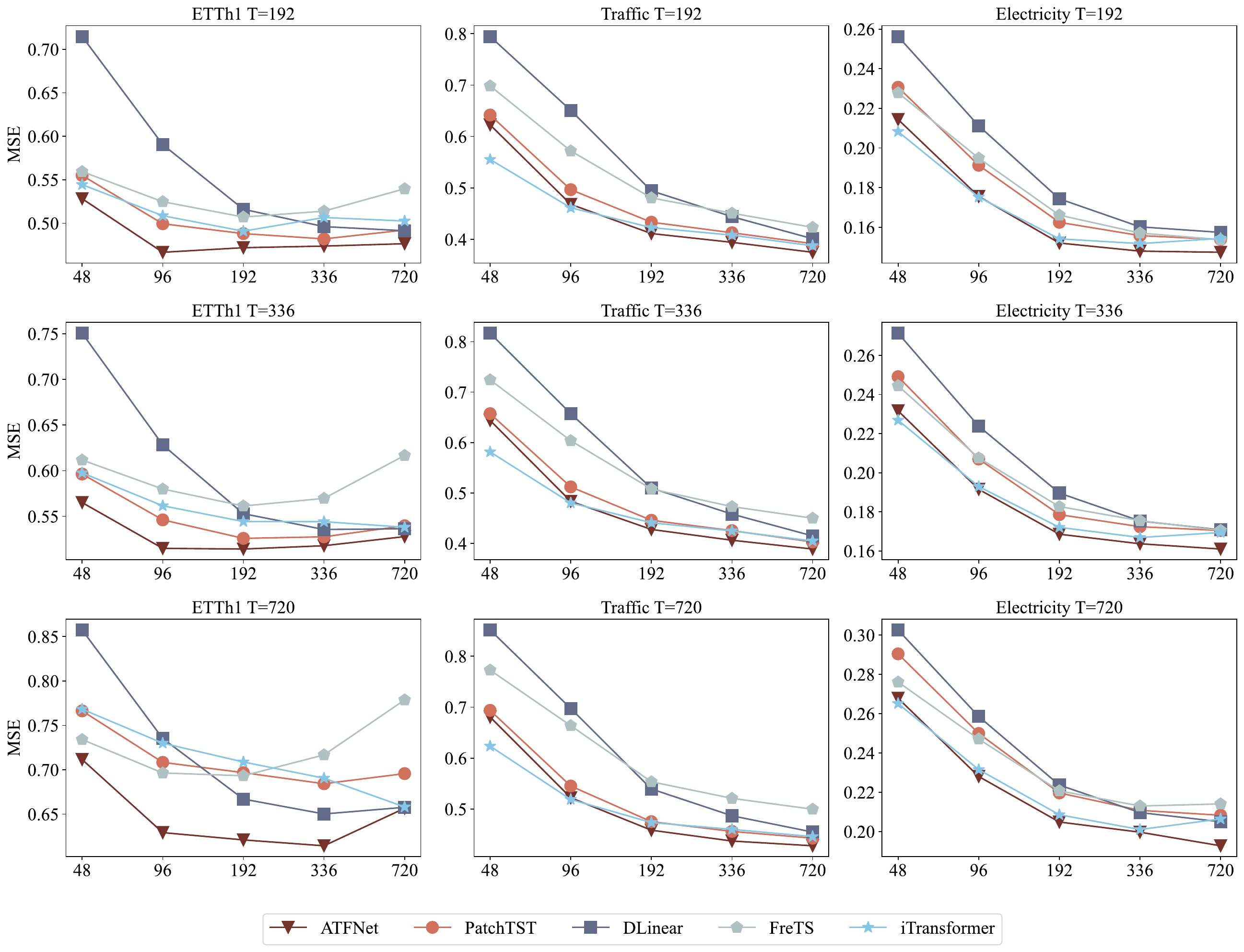}
\centering
\caption{Multi-variate long-term time series forecasting results (MSE) with 5 different look-back window size $L\in\{48, 96, 192, 336, 720\}$. The prediction horizon is $T\in \{192, 336, 720\}$.}
\label{full_lookback}
\end{figure}

\end{document}